\documentclass{article}
\usepackage{graphicx} 
\usepackage{amsmath}
\usepackage{amsfonts}
\usepackage{amssymb}
\usepackage{float}
\usepackage{url}
\usepackage[toc,page]{appendix}
\usepackage{tabularx}
\usepackage{longtable}
\usepackage{pdflscape}
\usepackage[margin=1in]{geometry}
\usepackage{color}
\usepackage{longtable}
\usepackage{lineno}

\usepackage{booktabs}
\usepackage{xcolor}
\usepackage{colortbl}

\usepackage[normalem]{ulem}

\newcommand{\code}[1]{\texttt{#1}}

\definecolor{honolulublue}{HTML}{0076B6}

\title{Universal Infectious Disease Forecasting with Mantis 2}
\author{Carson Dudley\footnote{Corresponding author, cdud@umich.edu}, Reiden Magdaleno, Yutong Bi, \\ Anthony Valencia, Xiaofeng Liu, Marisa Eisenberg}
\date{\today}

\begin{document}

\maketitle

\begin{abstract}

Public health agencies need to anticipate how outbreaks will unfold, when they will peak, and how long demand on health services will remain elevated. We previously developed a foundation model, Mantis, which was trained entirely on simulated outbreaks---enabling us to generate the diversity and scale of training data needed to support forecasting across diseases. However, our previous work left unresolved when simulation improves predictions and how it should complement observational data.

Here, we varied the balance of simulated and observed training data while holding forecasting tasks fixed and varying the available surveillance history. We found that simulation-rich training was particularly valuable with limited local history, and its relative advantage diminished as more observations became available. We also found that simulation-based training reduced exceptionally large forecast errors, suggesting that mechanistic knowledge can help compensate for limited local evidence.

Building on these findings, we developed Mantis 2, which combines simulated and observational training to jointly forecast up to six variables sixteen weeks ahead. Across 30 forecasting tasks, Mantis 2 achieved the best aggregate probabilistic accuracy over short, medium, and long horizons among the machine learning, statistical, and baseline models evaluated. It missed roughly 30\% fewer outbreak peaks than the strongest comparison model, with successfully anticipated peaks identified more than six weeks in advance on average. We publicly release Mantis 2, enabling public health agencies and researchers to forecast disease outcomes without developing a separate model for each disease or location.
\end{abstract}

\section*{Introduction}

Accurate infectious disease forecasts can inform hospital preparedness, staffing, treatment and vaccine allocation, risk communication, and public health interventions \cite{lutz2019path,desai2019realtime}. Their value is especially clear during emerging outbreaks, when decisions must be made rapidly despite incomplete information. Yet forecasting is particularly difficult in these settings: relevant historical data may be scarce, and changes in immunity, pathogen characteristics, behavior, or surveillance can make earlier observations less informative. Forecasting efforts for influenza, dengue, and COVID-19 have demonstrated useful short-term predictive performance, but accuracy varies substantially across diseases, seasons, and outbreak phases \cite{reich2019multiyear,johansson2019dengue,cramer2022covid}. Prominent operational systems have also generally concentrated on one- to four-week horizons, while expanding hospital capacity, procuring supplies, and planning vaccination campaigns can require longer preparation \cite{lutz2019path}. These challenges motivate forecasting systems that can be deployed rapidly across diseases and locations, draw on readily available surveillance signals, and provide useful predictions over longer planning periods.

Foundation models offer a way to reduce the need to develop a separate forecasting system for every task. Models such as Chronos, TimesFM, and Moirai learn from large collections of time series and transfer that knowledge to new forecasting targets without task-specific retraining \cite{ansari2024chronos,das2024timesfm,woo2024moirai}. Chronos-2 extends this capability to related variables and covariates \cite{ansari2025chronos2}. However, learning a broadly applicable forecasting model requires training examples that represent the range of settings it may encounter. For infectious diseases, surveillance records provide uneven coverage across pathogens and locations, and even extensive observational datasets often capture only a subset of possible outbreak dynamics (e.g., a certain type of disease, or in a certain location) \cite{bastos2019reporting,regimechange}.

Mechanistic simulation offers a way to broaden the training dataset. Models of transmission, immunity, interventions, and reporting can generate plausible outbreaks beyond those captured in surveillance records \cite{cobey2020dynamics,metcalf2020models}. Simulation-Grounded Neural Networks learn from these simulated trajectories, incorporating scientific knowledge through their training data \cite{sgnns}. The original Mantis demonstrated that a single model trained entirely on mechanistic simulations could forecast across diseases and surveillance settings \cite{mantis}. However, it forecast only one target at a time, with at most one covariate, and was limited to eight-week horizons. The exclusively simulated training curriculum also left unresolved how observation data and mechanistic knowledge could best complement one another.

Here we introduce Mantis 2, a general-purpose model that combines observational and simulated training to forecast infectious disease outcomes without disease-specific retraining. A single model jointly forecasts up to six variables sixteen weeks ahead, allowing signals such as cases, hospitalizations, wastewater, weather, and deaths to inform one another. It can use nearly ten years of historical observations when available, but can also generate forecasts with as little as 8 weeks of available data. These capabilities are intended to make forecasting useful with different levels of surveillance coverage and to provide a longer planning window without requiring users to develop or maintain multiple forecasting systems.

Mantis 2 also provides an opportunity to investigate when mechanistic training improves predictions beyond learning from observed outbreaks. Simulations can supply diverse epidemic trajectories, but their usefulness depends on how well they represent real disease dynamics and surveillance processes. Observational data capture these characteristics directly, but their coverage is uneven across diseases and locations \cite{bastos2019reporting}. A key distinction is between the information available during training and the local observations supplied when making a forecast. Even a model trained on many outbreaks may receive only a few weeks of history for a particular disease and location. We therefore ask whether simulated training is especially valuable when this local evidence is limited, and whether its contribution changes as surveillance history accumulates.

We evaluated Mantis 2 and a collection of comparison models across 30 disease-location forecasting tasks. It had the best aggregate probabilistic performance across short, medium, and long horizons. Across 3,725 observed outbreak peaks, Mantis 2 missed roughly 30\% fewer peaks than Chronos-2 (the second best model), with successful predictions providing more than six weeks of advance warning on average. To examine the contribution of simulation, we compare eleven training mixtures (with training data comprising more or less simulation data versus observational) while varying the historical context available. Simulation-rich training was particularly beneficial with limited context, its advantage diminished with longer histories, and it reduced exceptionally large forecast errors. These findings identify available surveillance history as an important determinant of the value of mechanistic training. By releasing Mantis 2, we aim to make these forecasting capabilities accessible to public health agencies and researchers without requiring a separate model for each disease or location.

Compared with the original Mantis, Mantis 2 extends the maximum forecast horizon from eight to sixteen weeks, increases usable historical context from 112 to 512 weeks, and supports joint forecasting of up to six variables rather than one target with an optional covariate. It combines observational and simulated training and replaces four separately deployed model variants with a single model. These expanded capabilities accompany substantial accuracy gains: in a separate comparison restricted to eight-week, single-target forecasting, Mantis 2 reduced pooled WIS by 31.42\% and improved performance on all 27 eligible comparison datasets. Full details of the comparison are provided in Appendix \ref{appendix:mantis_comparison}. Pretrained models, example data, evaluation scripts, and an interactive forecasting demo are available at {\texttt{www.github.com/carsondudley1/Mantis}}.

\begin{figure}[t]
    \centering
     \includegraphics[width=\textwidth]{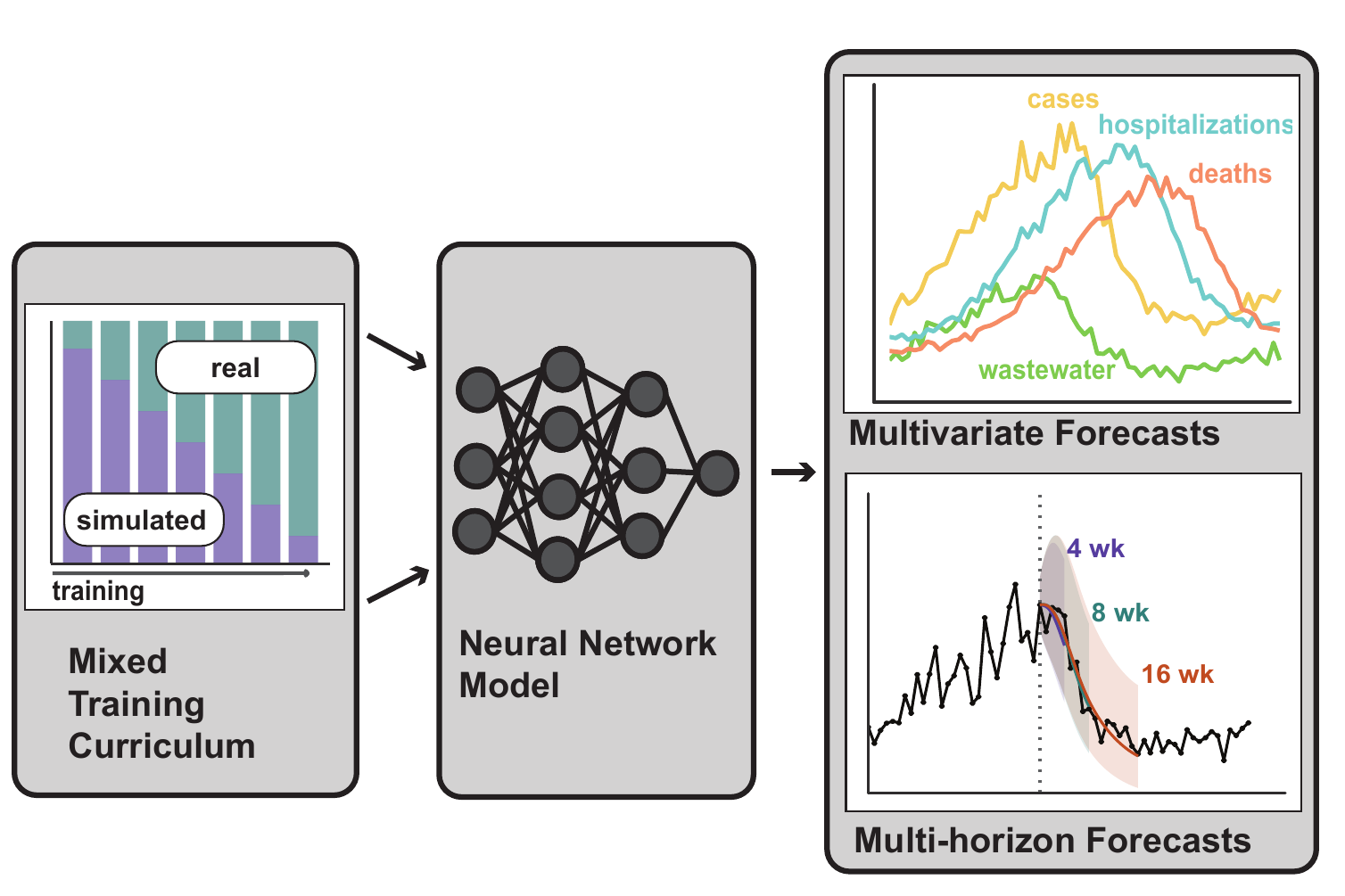}
 \caption{\textbf{Mantis 2 Workflow and Capabilities.} Mantis 2 builds upon the original Mantis foundation model for probabilistic infectious disease forecasting. In Mantis 2, the neural network training pipeline incorporates a mixed training curriculum that incorporates both real-world and simulated disease trajectories. Mantis 2 also brings two new capabilities: multivariate forecasting increased to 6 variables and multi-horizon forecasts up to 16 weeks.} 
    \label{fig:conceptual}
\end{figure}

\section*{Results}

Mantis 2 generates probabilistic forecasts across diseases, locations, and surveillance variables using a single trained model (Figure \ref{fig:conceptual}). We trained the model on over 1,000 real-world infectious disease scenarios and 20,000 mechanistically simulated scenarios (with the balance of real versus simulated data used for training determined by the curriculum described in the Methods section). Both simulated and real scenarios contain between one and six variables, with a combination of disease surveillance signals and relevant covariates (e.g., weather). Mantis 2 uses up to 512 weeks of historical observations to produce multivariate forecasts for the next 16 weeks.

\subsection*{Forecasting performance across tasks and horizons}

We evaluated Mantis 2 and all comparison models on 30 disease-location forecasting tasks, each potentially containing multiple time series (Table \ref{tab:eval_diseases}). These tasks include both held-out periods or locations for diseases represented in training, as well as diseases entirely absent from the training set. Each model we evaluated produced forecasts every four weeks, beginning after eight weeks of observations and using all available history up to a rolling maximum of 512 weeks. Each forecast covered the subsequent 16 weeks, so there is some overlap in evaluation windows.

We compared Mantis 2 forecasting performance to six baseline models: Chronos-2 \cite{ansari2025chronos2}, a three-layer long short-term memory network (LSTM) \cite{LSTM}, exponential smoothing (ETS) \cite{box2015time}, seasonal autoregressive integrated moving average (SARIMA) \cite{box2015time}, seasonal-naive forecasts (i.e., predict for a given week the value from the same time the previous year), and persistence forecasts (predict for each week in the forecast the last value before the forecast began). Every model received the same input data at each forecast point. The baseline models were trained separately for each disease, pooling observations across locations, whereas Mantis 2 and Chronos-2 were not modified for the evaluations. The baseline models requiring fitting were refit every 16 weeks.

\begin{figure}[htbp]
    \centering
     \includegraphics[width=\textwidth,  height=\textheight, keepaspectratio]{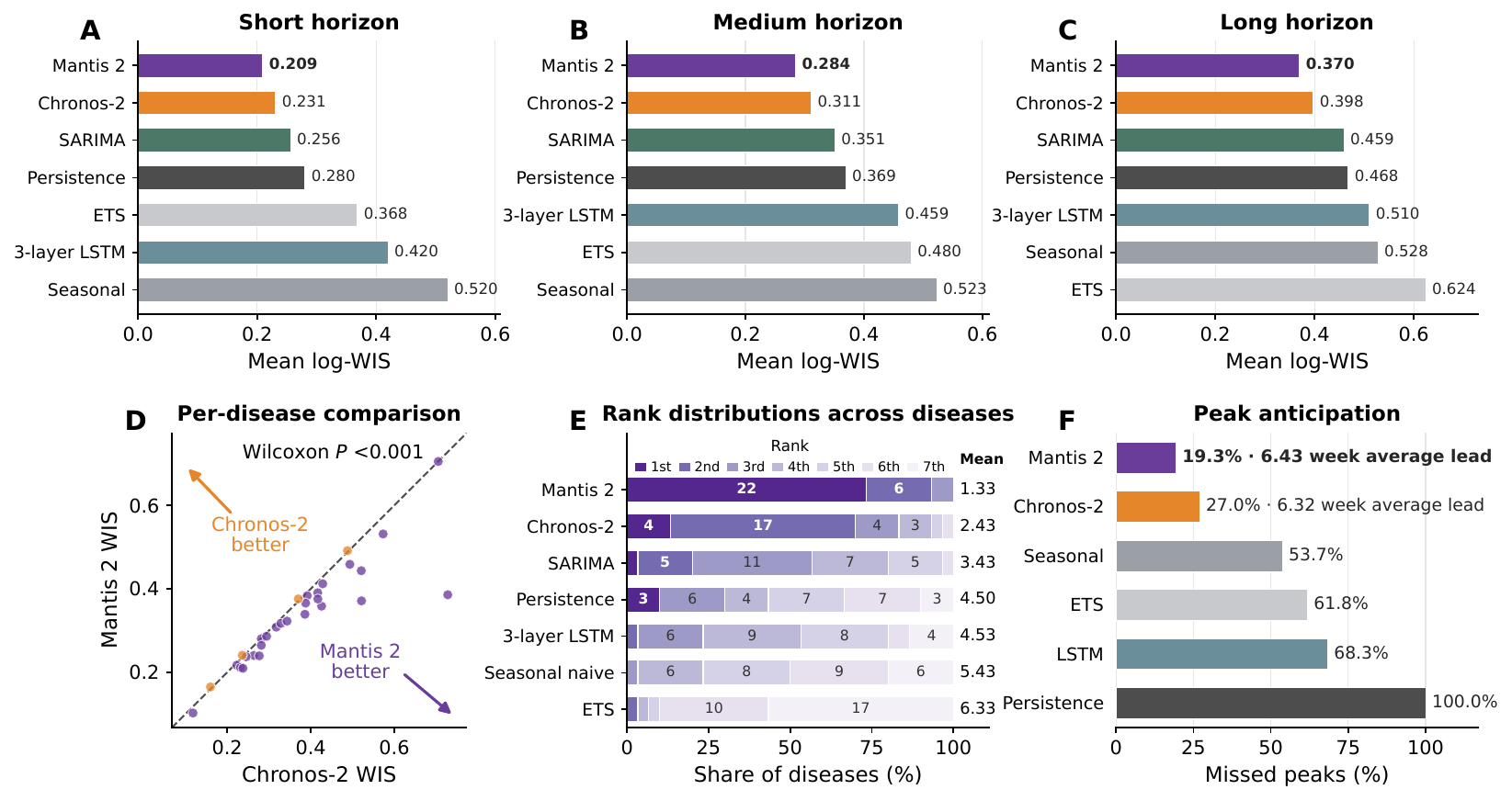}
     \caption{
    \textbf{Mantis 2 achieves strong probabilistic forecasting performance across diseases and forecast horizons.}
    \textbf{(A-C)} Mean weighted interval score evaluated on $\log(1+y)$-transformed outcomes (log-WIS) over short (1-4 weeks), medium (1-8 weeks), and long (1-16 weeks) forecast horizons. Lower scores indicate better performance. Mantis 2 achieves the lowest score across all three horizon ranges.
    \textbf{(D)} Disease-level comparison of Mantis 2 and Chronos-2 zero-shot over the long horizon. Each point represents an evaluation dataset, and the dashed diagonal indicates equal performance. Points below the diagonal favor Mantis 2 (purple), while points above favor Chronos-2 (orange). The paired Wilcoxon signed-rank test indicates a significant difference in performance ($P<0.001$).
    \textbf{(E)} Distribution of model ranks across the 29 evaluation datasets with results available for all six models, ranked by long-horizon WIS. Segment widths show the percentage of datasets at each rank. Numbers within segments indicate dataset counts, and mean ranks are shown on the right. Mantis 2 ranks first on 22 datasets and second on the remaining seven.
    \textbf{(F)} Peak prediction accuracy across 3,725 peaks. Bars show the percentage of outbreak peaks each model never correctly predicted at weekly lead times of 1--14 weeks (lower is better). A prediction is considered correct when the forecasted peak occurs within $\pm 2$ weeks of the observed peak. For Mantis 2 and Chronos-2 we report the mean lead time of the first correct peak prediction.
    }
    \label{fig:main_results}
\end{figure}

We evaluated the original Mantis separately because it supports forecasts only up to eight weeks ahead and predicts a single outcome with at most one covariate \cite{mantis}. It therefore could not participate in the full sixteen-week, joint multivariate evaluation summarized in Figure \ref{fig:main_results}. To compare the two versions within the original model's forecasting scope, we conducted a separate evaluation of single-target forecasts over weeks 1--8, reported in Appendix \ref{appendix:mantis_comparison}.

Probabilistic forecast accuracy was assessed using the weighted interval score (WIS) \cite{wis}, which rewards narrow predictive intervals while penalizing intervals that do not contain the observed outcome. We calculated WIS after applying the $\log(1+y)$ transformation to observations and forecast quantiles, reducing the influence of differences in outcome magnitude. Scores were summarized over short (1-4 week), medium (1-8 week), and long (1-16 week) forecast horizons. We also evaluated prediction interval calibration by comparing empirical coverage with target coverage levels (e.g., the 95\% prediction interval should contain the true outcome 95\% of the time). Finally, we assessed whether models correctly anticipated outbreak peak timing and how far in advance they did so. Anticipating peaks requires predicting a turning point before the decline is observed, making it a particularly challenging forecasting task. Advance warning of these turning points can support hospital staffing, capacity planning, and the allocation of supplies during periods of greatest demand.

Mantis 2 achieved the lowest aggregate log-WIS across all models for all three horizon ranges (Figure \ref{fig:main_results}A-C). Performance gains were distributed across evaluation tasks. Mantis 2 ranked first among all evaluated models the majority of the time, with average rank 1.33. Since Chronos-2 was consistently the closest model to Mantis 2's performance, we evaluated whether the performance gain from Chronos-2 to Mantis 2 was statistically significant. We did a paired Wilcoxon signed-rank test, using the performance of each model on each dataset, and found that Mantis 2's performance gain was statistically significant ($P<0.001$, Figure \ref{fig:main_results}D). Task-level rankings provide a complementary assessment to the forecast-weighted aggregate scores by describing how consistently each model performs across distinct forecasting settings (Figure \ref{fig:main_results}E).

Mantis 2 also substantially improved on its predecessor within this more restricted forecasting setting. Across 27 eligible datasets and 34,762 matched forecast origins, it reduced pooled eight-week WIS by 31.42\% relative to original Mantis, with improvements on every dataset. Dataset-level reductions ranged from 10.37\% to 45.55\% (Appendix \ref{appendix:mantis_comparison}). Both models forecast the same targets at the same origins, while retaining their respective historical context lengths and auxiliary-input configurations. Thus, Mantis 2 improves accuracy on tasks supported by the original model while also extending forecast horizons and enabling joint forecasting of multiple outcomes.

\begin{figure}
    \centering
     \includegraphics[width=\textwidth,  height=\textheight, keepaspectratio]{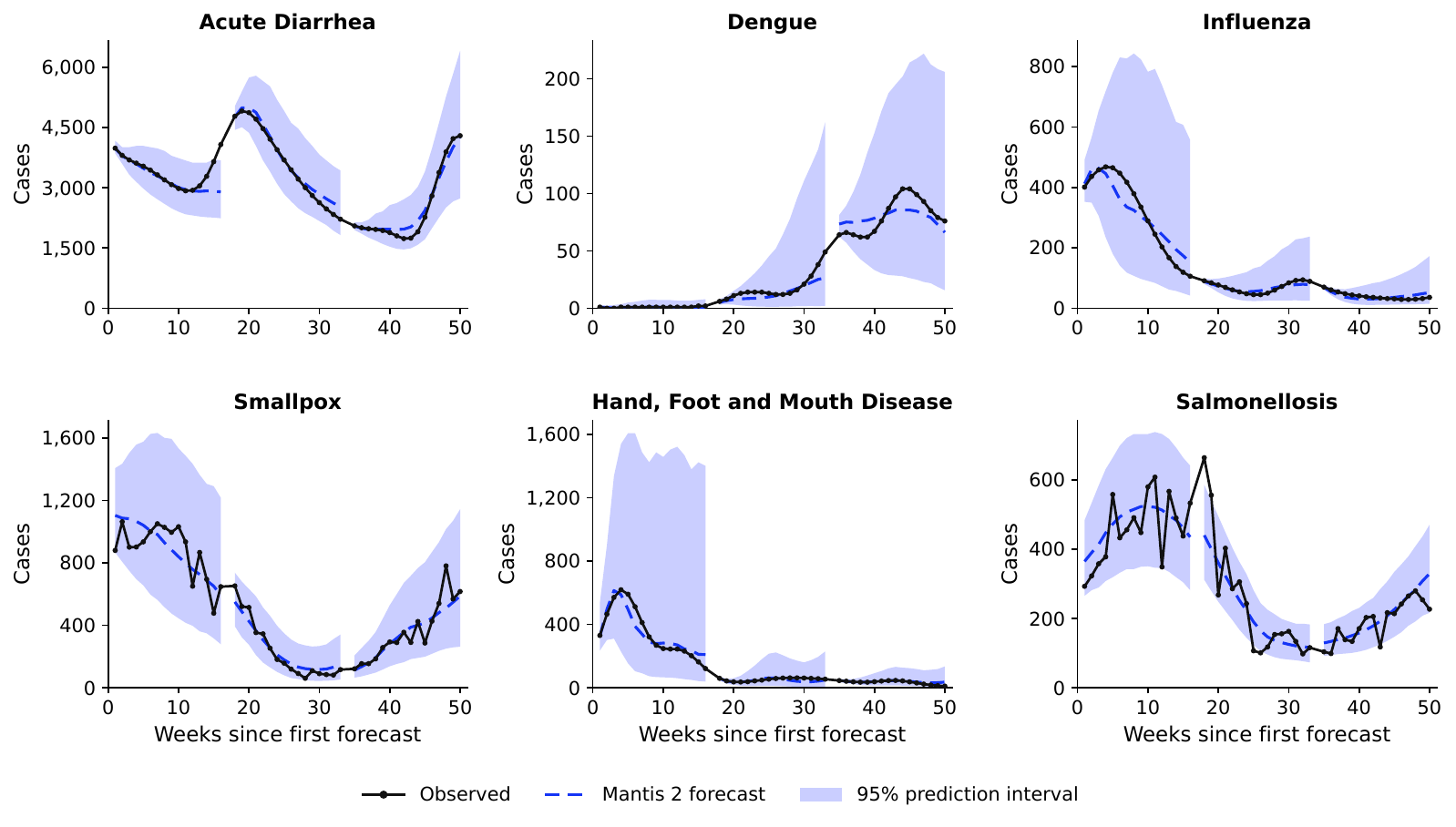}
    \caption{\textbf{Forecasts from Mantis 2 over a year of six diseases.} We display 16-week forecasts with a one week gap between, covering 51 weeks of each disease. The forecasts are generally accurate, with higher uncertainty (wider prediction intervals) during outbreak periods.}
    \label{fig:mantis_forecasts}
\end{figure}

Figure \ref{fig:mantis_forecasts} illustrates forecasts across six disease and location settings, from modern disease forecasts to historical epidemics. Each panel displays three 16-week forecasts issued 17 weeks apart. The panels therefore show forecasting over approximately one year, allowing predictions to be shown across seasonal rises, peaks, and declines. Although the forecasts are multivariate, only cases are displayed to facilitate comparison across diseases. Forecast medians and 95\% prediction intervals illustrate both the predicted trajectories and their uncertainty.

\subsection*{Anticipating outbreak peaks}

Accurate forecasts of near-term disease activity do not necessarily imply that a model can anticipate when an outbreak will peak. A model can track an ongoing rise while failing to predict the transition from growth to decline. Anticipating this transition requires information about how an outbreak may develop beyond its current trajectory. It also addresses a practical planning question: how much longer will disease activity increase before it begins to subside? This timing can inform preparations for sustained demand on health services, including staffing and resource allocation. Because these preparations take time, a useful forecast must identify the turning point further in advance to support action.

We therefore conducted a separate evaluation of outbreak peak prediction across 3,725 observed peaks, assessing both whether models correctly predicted peak timing and how much advance warning they provided. Predicted peaks were identified as local maxima in the forecast median (0.5 quantile forecast), with two subsequent weeks of decline required to establish a peak. Thus, a 16-week forecast could identify peaks at lead times of up to 14 weeks. A peak prediction was considered correct when its timing fell within two weeks of the observed peak. We measured the proportion of observed peaks that were  correctly anticipated across lead times of 1-14 weeks and, among successfully anticipated peaks, the lead time of the first correct prediction.

Mantis 2 correctly predicted over 80\% of observed peaks, compared with 73\% for Chronos-2, 46.3\% for naive seasonal forecasts, 38.2\% for ETS, 31.7\% for the LSTM, and 0\% for naive persistence (Figure \ref{fig:main_results}F). Relative to Chronos-2, Mantis 2 reduced the proportion of missed peaks by roughly 30\%. Among peaks each model successfully anticipated, the mean lead time of the first correct prediction was 6.43 weeks for Mantis 2 and 6.32 weeks for Chronos-2. These averages are conditional on model-specific subsets of successfully anticipated peaks and therefore do not represent a paired comparison of warning time (if you anticipate fewer peaks, it stands to reason that the ones you do anticipate are more obvious, so your lead time may be inflated). Mantis 2  correctly anticipated a larger proportion of observed peaks, with successful predictions providing more than six weeks of advance warning on average.

\subsection*{Exploring the optimal balance of simulated and real world training data}

\begin{figure}[htbp]
    \centering
    \includegraphics[width=\textwidth]{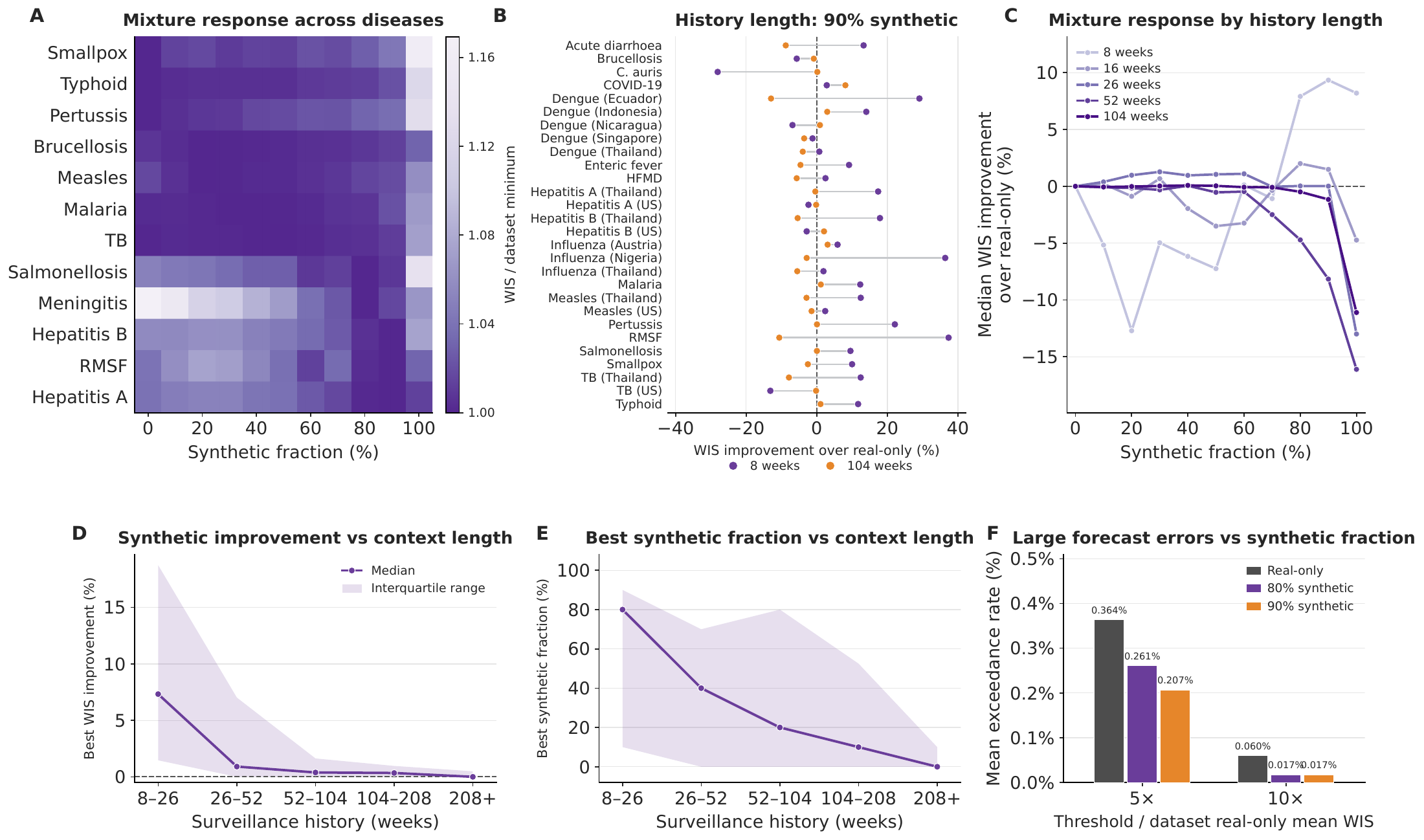}
    \caption{
    \textbf{Synthetic training benefits depend on the dataset and available history.}
    \textbf{(A)} WIS across synthetic training fractions for Project Tycho datasets, normalized by each dataset's minimum across mixtures. Darker cells indicate better performance. 
    \textbf{(B)} Improvement over real-only training for the 90\% synthetic model with 8 versus 104 weeks of history. Connected points represent the same dataset and forecast origins. Positive values favor high-synthetic training.
    \textbf{(C)} Median improvement over real-only across all synthetic mixtures at five context lengths.
    \textbf{(D)} Observed improvement of best synthetic mixture over real-only within each surveillance-history. Synthetic pretraining mixtures are more useful with shorter context length.
    \textbf{(E)} Synthetic fraction attaining the lowest WIS for each context length. In D-E, points and shading show medians and interquartile ranges across datasets.
    \textbf{(F)} Frequency of forecast WIS exceeding five or ten times each dataset's mean real-only WIS. Models with higher amounts of synthetic pretraining make extreme errors less frequently.
    }
    \label{fig:synthetic_blends}
\end{figure}

An important question for developing infectious disease foundation models is whether training should emphasize the outbreaks that have been observed or the broader range of outbreaks that could occur. Synthetic pretraining offers a way to learn from the latter, since you can simulate a very broad range of disease dynamics. Strong predictive performance has been achieved across domains by models pretrained entirely on synthetic datasets \cite{tabpfn,qu2025tabicl}, and the original Mantis demonstrated that training exclusively on mechanistic simulations could support forecasting across real-world diseases and surveillance settings \cite{mantis}. Simulations can expand training beyond the (comparatively) limited number of outbreaks that have been observed and have high-quality, publicly available datasets and encode explicit scientific assumptions about transmission and observation. These assumptions provide an interpretable basis for the training distribution, and a model trained on these simulations can make predictions that are highly interpretable \cite{sgnntheory}, but the utility of synthetic pretraining depends on how well the simulations represent the processes encountered in practice.

The simulation ensemble we trained on, Reservoir \cite{reservoir}, spans human, vector, and environmental transmission, including asymptomatic infection, interventions such as quarantine and isolation, and pre- and post-exposure prophylaxis. It also represents the observation process through underreporting, reporting delays, and other surveillance effects. Because the simulator is broad and realistic, one might hypothesize that simulated training would provide sufficient diversity for general forecasting, with observational data contributing little additional information. On the other hand, even a simulator that can represent a broad range of diseases in a realistic manner could omit important dynamics, misrepresent their frequency, or not accurately represent the noise and observation structure of real-world infectious disease surveillance. Observational training could supply these missing patterns and, if sufficiently diverse, make simulated training redundant.

Observational data introduce a different concern: the datasets we have access to may not represent those in which forecasts are most needed. For example, our training dataset includes a lot of respiratory surveillance data obtained through Delphi Epidata \cite{delphi_epidata}, whose COVID-19 and flu data provide extensive coverage of respiratory diseases in the United States. Heavy reliance on these datasets could teach the model to expect dynamics specific to these respiratory diseases in the United States, but hurt performance in less-represented settings (for example, one of our evaluation datasets: dengue in Asian and South American countries). This creates a potential health equity concern if forecasting benefits depend on the availability of extensive historical surveillance.

We therefore compared models trained with fixed synthetic fractions from 0\% to 100\% (i.e., from training only on real-world data, to training only on simulated data, and with mixtures in between) in increments of 10 percentage points across a range of infectious disease types and with locations around the world. We only measure the performance of fixed mixtures of synthetic versus real data here; our released Mantis 2 uses a training curriculum that yields better results than any of the fixed quantities shown (see Methods and Appendix for more details on Mantis 2 training curriculum).

Neither observational nor simulation-rich training was uniformly preferred across datasets (Figure \ref{fig:synthetic_blends}A). Within Project Tycho data \cite{project_tycho} (historical US disease records), for example, smallpox and pertussis favored observational training, whereas hepatitis A and B favored simulation-rich mixtures. Malaria and tuberculosis showed relatively little variation across a broad range of mixtures. These responses raised a further possibility: the usefulness of simulated training might depend not only on which disease is forecast, or the location in question, but also on how much local evidence the model receives.

To evaluate the impact of surveillance history for a given disease on the amount of synthetic data that was useful, we varied the available context while keeping the forecasting task fixed. We evaluated all synthetic mixtures with 8, 16, 26, 52, and 104 weeks of history at each forecast origin across all evaluation datasets. With only eight weeks of observations, the 80\% and 90\% synthetic models improved median WIS over real-only training by 7.9\% and 9.3\%, respectively. With 104 weeks, their median improvements were instead $-0.5\%$ and $-1.2\%$. The relative advantage of 90\% synthetic training was greater with short history in the vast majority of datasets (Figure \ref{fig:synthetic_blends}B). Thus, simulation-rich training was particularly useful when the model had little local evidence on which to condition its forecast. Simulation-only training improved median WIS by 8.2\% with eight weeks but worsened it by 11.1\% with 104 weeks (Figure \ref{fig:synthetic_blends}C). Our results do not support the notion that either synthetic nor observation data are alone sufficient to train a state-of-the-art foundation model for infectious disease. Instead, they show an interaction between the training distribution and the information supplied at prediction time.

That interaction also changed apparent disease-specific preferences. Although smallpox and pertussis favored observational training in the original evaluation, the 80\% synthetic model improved their WIS by 8.8\% and 18.3\%, respectively, with only eight weeks of context. With 104 weeks, the corresponding improvements were $-1.2\%$ and $-0.5\%$. A preference for observational training therefore cannot, by itself, be interpreted as evidence that a disease is poorly represented by the simulator. A rolling-origin evaluation showed the same pattern: both the median benefit of the best mixture of real and synthetic data declined as surveillance histories lengthened (Figure \ref{fig:synthetic_blends}D,E).

We also examined whether training composition affected exceptionally large forecast errors, which can be obscured by summaries of typical performance. Across our evaluation datasets datasets, the mean fraction of forecasts that had error greater than five times the dataset's mean real-only WIS was 0.364\% for real-only training, compared with 0.261\% and 0.207\% for 80\% and 90\% synthetic training. At ten times the reference mean, the rates were 0.060\%, 0.017\%, and 0.017\%, respectively (Figure \ref{fig:synthetic_blends}F). Simulation-rich training could therefore mitigate extreme failures even in cases where it provides perhaps less overall performance benefit due to the availability of more extensive real-world data.

These findings suggest that the contribution of simulated training depends on the evidence available to the model. Broad mechanistic training was particularly valuable when context with a specific disease was very limited, whereas longer histories often favored observational training or reduced sensitivity to the mixture. It seems that a simulated prior can provide a model with knowledge of the full range of plausible infectious disease outcomes and their relative probabilities, which is very useful with limited context, and which helps the model avoid catastrophic errors due to predicting improbable outcomes. However, once the model has an established history of a given disease, this scientific prior is not as useful, because the model is very skilled at adapting its predictions to a real-world disease history.

\section*{Discussion}

Mantis 2 provides a common forecasting model for infectious diseases across pathogens, locations, and surveillance signals without disease-specific retraining. Its multivariate forecasts extend up to sixteen weeks ahead, and its strong performance across the evaluated tasks demonstrates the potential for a shared model to support forecasting in diverse epidemiological settings. Mantis 2 also anticipated a larger proportion of observed outbreak peaks than the evaluated baselines, with successful predictions providing more than six weeks of advance warning on average. Together, these capabilities expand the range of forecasting tasks that can be addressed using a single pretrained system.

For public health agencies and researchers, a general-purpose model can reduce the need to develop and maintain a separate forecasting system for each disease and surveillance setting. Mantis 2 can jointly use multiple surveillance variables, allowing related signals to inform predictions while also accommodating settings in which only a single series is available. Its longer forecast horizon is especially useful because many important decisions  require weeks of lead time, including staffing, procurement, and capacity planning. Peak forecasts provide complementary information about when disease activity may decrease and how long elevated demand may persist. These capabilities could make forecasts useful for a broader range of planning decisions.

Our training experiments identify one very relevant consideration for extending forecasting to settings with limited data: the value of mechanistic simulation depends on the amount of local surveillance history available at prediction time. Simulation-rich training was especially beneficial with short histories, and even diseases that favored observational training with longer histories benefited from greater simulated exposure when context was limited. Mechanistic training may therefore supply useful background knowledge (scientific prior) before a surveillance series has accumulated enough observations to establish its own recurring patterns. For emerging outbreaks and newly established surveillance streams, the practical implication is that useful forecasting need not depend entirely on having years of local observations. An interesting future work would be investigating how the model internally combines simulated and observational information.

Our mixture experiments held the proportion of simulated training examples constant throughout training, whereas the released Mantis 2 uses a curriculum that gradually reduces this proportion from 70\% to zero. The mixture experiments therefore clarify how training composition affects forecasting, but do not establish how best to schedule exposure to simulated and observational data. Future work could repeat these comparisons under different curricula to examine how the amount and timing of simulated training jointly affect performance, particularly when local surveillance history is limited.

Forecasting reliability also matters beyond average accuracy. Simulation-rich training reduced the frequency of exceptionally large forecast errors, suggesting that training composition can influence the risk of occasional extreme failures as well as typical performance. Such failures are particularly relevant to operational use, where an implausible forecast can undermine confidence or distort preparations. This result supports evaluating foundation models across multiple dimensions, including average probabilistic accuracy, anticipation of turning points, and the frequency of extreme errors. Further work should establish which failure modes mechanistic training reduces and whether these benefits persist in prospective forecasting. 

Several limitations qualify these findings. Our evaluation of Mantis 2 is retrospective, and improved forecast scores do not by themselves demonstrate improved public health decisions. However, the original Mantis has shown prospective use \cite{utah}, and Mantis 2 outperforms its predecessor in our retrospective comparison by more than 30\% (Appendix \ref{appendix:mantis_comparison}). Likewise, the ability to generate sixteen-week forecasts does not imply equal reliability across horizons or settings. Both the observational training data and the mechanistic simulations represent selected diseases, locations, and processes; their coverage cannot guarantee performance for every emerging pathogen or surveillance system. Finally, our conclusions about how surveillance history affects the value of simulated training are based on experiments that truncate existing series while holding forecast origins fixed. These experiments isolate the effect of having fewer observations, but a short history drawn from an established series does not necessarily represent the dynamics or uncertainty encountered early in a new outbreak. We therefore hypothesize that simulation-rich training will be particularly useful early in emerging outbreaks, but do not directly test that hypothesis here. Additionally---the fact the optimal synthetic fraction varies across diseases and context lengths raises the question of what blend may be best to use for a multi-purpose model such as Mantis 2, which is intended as a foundation model to be used across settings. In this work, we opted to use a blend that amount to 35\% synthetic data, but future work would be useful to explore further optimizing this mix (balancing across diseases and between the goals of overall performance vs. minimizing large errors), as well as to evaluate which simulations (models, real-world effects, etc.) most improve performance.

Mantis is already being used by several U.S. state governments and is supporting forecasting efforts by federal governments on other continents. This existing use provides a practical setting for evaluating and extending the capabilities introduced in Mantis 2. By releasing the model, we aim to make multivariate, longer-horizon disease forecasting more accessible to public health agencies and researchers without requiring each group to train its own disease-specific system. The release will also enable independent evaluation across additional pathogens and surveillance settings. Mantis 2 offers a foundation for expanding access to infectious disease forecasts, including where limited local history makes forecasting both difficult and urgently needed.

\section*{Methods}

\subsection*{Training data and preprocessing}

\begin{longtable}{
  @{}
  >{\raggedright\arraybackslash}p{4.cm}
  >{\raggedright\arraybackslash}p{3.cm}
  >{\raggedright\arraybackslash}p{4.1cm}
  >{\raggedright\arraybackslash}p{4.cm}
  @{}
}
\caption{Evaluation Dataset. The evaluation dataset for Mantis 2 consists of 23 diseases across 11 countries and eight transmission categories (enteric,
vector-borne, respiratory, airborne, bloodborne, close contact, zoonotic, and
healthcare-associated). Most datasets
consist of univariate case counts; multivariate datasets include additional surveillance
signals.} \label{tab:eval_diseases} \\
\toprule
\multicolumn{4}{l}{\textit{In-domain evaluation}} \\
\toprule
\textbf{Disease} & \textbf{Location} & \textbf{Transmission} & \textbf{Signals} \\
\midrule
\endfirsthead
\toprule
\textbf{Disease} & \textbf{Location} & \textbf{Transmission} & \textbf{Signals} \\
\midrule
\endhead
\midrule
\multicolumn{4}{r}{\textit{Continued on next page}} \\
\endfoot
\bottomrule
\endlastfoot
Acute diarrhea \cite{Thai_df} & Thailand (Provincial) & Enteric & Univariate \\
Enteric fever \cite{Thai_df} & Thailand (Regional) & Enteric & Univariate \\
Hepatitis A \cite{Thai_df} & Thailand (Regional) & Enteric & Univariate \\
Typhoid fever \cite{van_panhuis_2018_12608994} & USA (Regional) & Enteric & Univariate \\
Salmonellosis \cite{van_panhuis_2018_12608994}& USA (Regional) & Enteric & Univariate \\
Hepatitis A \cite{van_panhuis_2018_12608994} & USA (State) & Enteric & Univariate \\
\midrule
Dengue \cite{clarke2024global} & Singapore (National) & Vector-borne & Univariate \\
Dengue \cite{clarke2024global} & Nicaragua (Provincial) & Vector-borne & Univariate \\
Dengue \cite{clarke2024global}& Indonesia & Vector-borne & Univariate \\
Dengue \cite{clarke2024global} & Thailand (Provincial)  & Vector-borne & Univariate \\
Malaria \cite{van_panhuis_2018_12608994} & USA (Regional) & Vector-borne & Univariate \\
Rocky Mountain Spotted Fever  \cite{van_panhuis_2018_12608994} & USA (National) & Vector-borne & Univariate \\
\midrule
Influenza \cite{Thai_df}  & Thailand (Provincial)  & Respiratory & Univariate \\
HFMD  \cite{Thai_df} & Thailand (Provincial)  & Respiratory, Enteroviral  & Univariate \\
Measles \cite{van_panhuis_2018_12608994} & USA (State) & Respiratory & Univariate \\
Whooping cough \cite{van_panhuis_2018_12608994} & USA (Regional) & Respiratory & Univariate \\
Influenza (ILI) & Nigeria & Respiratory & Multivariate (ILI, SARI, ARI, PNEU, deaths) \\
\midrule
Tuberculosis \cite{Thai_df} & Thailand (Provincial)  & Bloodborne, Airborne & Univariate \\
Hepatitis B \cite{Thai_df} & Thailand (Provincial)  & Bloodborne  & Univariate \\
Hepatitis B \cite{van_panhuis_2018_12608994} & USA (Regional) & Bloodborne  & Univariate \\
Smallpox & USA (Regional) & Close contact & Univariate \\
Brucellosis \cite{van_panhuis_2018_12608994} & USA (Regional) & Zoonotic & Univariate \\
Tuberculosis \cite{van_panhuis_2018_12608994} & USA (State) & Bloodborne, Airborne & Univariate \\
\midrule
\multicolumn{4}{l}{\textit{Out-of-domain evaluation}} \\
\midrule
COVID-19 \cite{epidata} & USA (State) & Respiratory  & Multivariate (cases, deaths, hospitalizations, ED visits) \\
Candida auris \cite{cdc_nndss_weekly} & USA (National) & HAI  & Univariate \\
Carbapenem-resistant Enterobacteriaceae \cite{cdc_nndss_weekly} & USA (National) & HAI  & Univariate \\
Dengue \cite{clarke2024global} & Ecuador (National)  & Vector-borne  & Univariate \\
Meningitis \cite{van_panhuis_2018_12608994}& USA (Regional) & Respiratory, Close contact  & Univariate \\
Influenza (ILI) \cite{whoflunet} & Austria & Respiratory  & Multivariate (ILI, SARI, ARI, PNEU, deaths) \\
Measles  \cite{Thai_df} & Thailand (Regional) & Respiratory  & Univariate \\
Smallpox \cite{van_panhuis_2018_12608994}& USA (Regional) & Close contact & Univariate \\
Mpox \cite{who_mpox_global_2024} & Sierra Leone & Close contact, Zoonotic & Multivariate (cases, deaths) \\
Ebola virus disease \cite{moh_drc_ebola_2020}& DRC (National) & Close contact,  Zoonotic & Multivariate (confirmed cases, deaths, probable cases) \\
Cholera \cite{mosaic} & Ethiopia (National) & Enteric & Multivariate (cases, deaths) \\
\end{longtable}

Mantis 2 was trained on observational infectious disease surveillance data and epidemic trajectories generated using the Reservoir mechanistic simulator \cite{reservoir}. Observational data covered multiple diseases, locations, and surveillance systems, with individual examples containing up to six variables. Approximately 20,000 simulated scenarios represented human, vector, and environmental transmission under varied epidemiological parameters, population characteristics, interventions, and observation processes, including underreporting and reporting delays. Both sources supplied multivariate combinations of surveillance signals. Full data sources, inclusion criteria, and simulator settings are provided in the Appendix.

Time series were represented at weekly resolution, with counts aggregated to this resolution if daily values were available. Missing observations were infrequent and were filled using spline interpolation over each complete series before temporal splitting. Imputed values were retained in evaluation targets and included in scoring. Inputs were subsequently transformed using the inverse hyperbolic sine function as a variance-stabilizing transformation \cite{ansari2025chronos2} and normalized using instance normalization \cite{revin}. Complete preprocessing and training-example construction procedures are described in the Appendix.

\subsection*{Model architecture and training}

Mantis 2 was initialized from pretrained Chronos-2 weights \cite{ansari2025chronos2}, and the full model was trained on the combined infectious disease data. The model accepts up to six variables and uses up to 512 weeks of historical context to generate probabilistic forecasts for the subsequent sixteen weeks. Calendar covariates provide additional seasonal information. Forecasts are represented by predicted quantiles for each target and forecast horizon, with forecasts provided for 21 quantiles: [0.01, 0.05, 0.10, 0.15, 0.20, 0.25, 0.30,
 0.35, 0.40, 0.45, 0.50, 0.55, 0.60, 0.65,
 0.70, 0.75, 0.80, 0.85, 0.90, 0.95, 0.99]. We incorporated latent chain-of-thought computation with two recurrences, allowing intermediate hidden representations to undergo additional processing before forecast generation.

Training used a curriculum in which the probability of sampling simulated examples decreased linearly from 70\% to zero. This schedule progressively shifted training from predominantly simulated trajectories to entirely observational data. We used the final training checkpoint for evaluation, without early stopping. The architecture, training objective, optimization settings, and computational requirements are detailed in the Appendix.

\subsection*{Evaluation datasets and forecast protocol}

Evaluation used two forms of data separation. For disease-location datasets represented in training, we used a temporal split. For out-of-domain evaluation, entire disease-location datasets were excluded from Mantis 2 training. Here, out-of-domain refers to an unseen dataset rather than necessarily an unseen disease: a disease could be represented in training through data from another location. Evaluation datasets and their surveillance signals are listed in Table \ref{tab:eval_diseases}.

Forecasts were issued at rolling origins spaced four weeks apart, with a minimum of eight weeks of historical context. 
At each origin, models received the preceding time series, subject to a maximum context length of 512 weeks, and forecast the following sixteen weeks. Designated surveillance targets contributed to evaluation scores, while auxiliary variables (e.g., weather) could inform forecasts without being scored. Mantis 2 and Chronos-2 were applied without task-specific parameter updates.

We compared Mantis 2 with Chronos-2 \cite{ansari2025chronos2}, as well as a three-layer long short-term memory (LSTM) neural network \cite{LSTM}, exponential smoothing (ETS), seasonal-naive forecasts, seasonal autoregressive integrated moving average (SARIMA), and persistence forecasts. Seasonal-naive forecasts used the corresponding observation from the previous year, while persistence forecasts used the most recent value as their central prediction. Baseline fitting procedures, refitting schedules, and uncertainty estimation are described in the Appendix, together with an additional comparison against the original Mantis.

\subsection*{Forecast accuracy and peak prediction}

Probabilistic forecast accuracy was assessed using the weighted interval score (WIS) \cite{wis}, with lower values indicating better performance. WIS was calculated after applying the $\log(1+y)$ transformation to observations and forecast quantiles. We summarized performance over short (weeks 1-4), medium (weeks 1-8), and long (weeks 1-16) horizons. Models were additionally ranked within evaluation datasets, and Mantis 2 and Chronos-2 were compared using a paired Wilcoxon signed-rank test on dataset-level long-horizon WIS. Metric definitions, score aggregation, and treatment of unavailable predictions are specified in the Appendix.

We conducted a separate peak-timing evaluation using forecasts issued at weekly origins. Observed peaks were identified using the detection and eligibility criteria described in the Appendix. Predicted peaks were identified as local maxima in the forecast median, requiring two subsequent weeks of decline. A sixteen-week forecast could therefore identify peaks at lead times of up to fourteen weeks. A prediction was considered correctly timed when the predicted peak fell within two weeks of the observed peak.

For each model, we calculated the proportion of observed peaks never correctly anticipated at lead times of one to fourteen weeks. Among successfully anticipated peaks, we calculated the mean lead time of the first correct prediction. These mean lead times were conditional on model-specific subsets of successfully anticipated peaks and therefore did not constitute a paired comparison of warning time.

\subsection*{Training-mixture and surveillance-history experiments}

To examine the contribution of simulated training data, we trained eleven models using constant simulated fractions from 0\% to 100\% in increments of ten percentage points, holding the remaining training configuration fixed. These models were distinct from the curriculum-trained Mantis 2 model.

We evaluated each mixture with 8, 16, 26, 52, and 104 weeks of historical context, holding forecast origins and scored outcomes fixed across context lengths. This design isolated the effect of the amount of history supplied at prediction time. For each dataset and context length, improvement relative to real-only training was calculated as
\begin{equation*}
100 \times \frac{\mathrm{WIS}*{\mathrm{real}}-\mathrm{WIS}*{\mathrm{mixture}}}
{\mathrm{WIS}_{\mathrm{real}}},
\end{equation*}

so that positive values indicated an advantage from simulated training. We summarized improvements across datasets and identified the simulated fraction attaining the lowest WIS at each context length. We also examined mixture performance as available surveillance history accumulated during rolling-origin evaluation.

To assess extreme forecast errors, we calculated the frequency with which forecast WIS exceeded five or ten times the corresponding dataset's mean real-only WIS. The same dataset-specific thresholds were applied across training mixtures.

\section*{Acknowledgments}
This project was made possible by the Insight Net cooperative agreement with University of Michigan (5 NU38FT000002-02-00) from the CDC’s Center for Forecasting and Outbreak Analytics (CDC-RFA-FT-23-0069). Its contents are solely the responsibility of the authors and do not necessarily represent the official views of the Centers for Disease Control and Prevention. The authors thank Julie Gilbert for her assistance with gathering data for an earlier version of the manuscript. The authors thank Kevin M. Bakker for his assistance in finding and collecting data from the Thai Ministry of Public Health. The authors also thank the two anonymous reviewers whose comments have greatly strengthened the manuscript.

\bibliographystyle{unsrt}
\bibliography{references}

\begin{appendices}

\section{Differences from the original Mantis}

Mantis 2 retains the original Mantis' objective of providing probabilistic infectious disease forecasts across pathogens, locations, and surveillance settings without disease-specific retraining \cite{mantis}. Its development addresses several practical limitations of the original model: the amount of historical information it could use, the number of surveillance variables it could incorporate and forecast, the length of its forecast horizon, and the need to select among separate model variants, depending on the forecast horizon and if multiple variables are used. We also broadened the training approach to combine mechanistic simulations with observational data. Below, we describe these changes and the considerations that motivated them.

\subsection*{Longer historical context and a more efficient architecture}

We increased the maximum historical context from 112 to 512 weeks, extending the usable surveillance history from just over two years to nearly ten years. A two-year window provides relatively few examples of seasonal recurrence and may omit less frequent but informative epidemic patterns. For example, a recent history containing only single-peaked influenza seasons may provide limited local evidence for anticipating a season with two distinct peaks. A longer window can include previous instances of such behavior, as well as  greater variation in epidemic timing, magnitude, and duration. These observations can help the model interpret an ongoing outbreak in the context of a broader history of disease activity. Our surveillance-history experiments support the value of additional context over the lengths evaluated. The expanded context capacity allows longer records to be used when available.

Increasing context length required an architecture that could process longer sequences efficiently. Our most important design consideration for both Mantis models (beyond that they forecast accurately) is practical accessibility: generating forecasts should be feasible on a laptop, with computational requirements compatible with routine use by public health practitioners and researchers. Simply extending the sequence length of the original custom CNN-transformer architecture we used for the original Mantis would increase the computational burden of temporal attention (which scales quadratically in sequence length, so making the context 5x longer would cost 25x as much). We therefore adopted the Chronos-2 backbone \cite{ansari2025chronos2}, whose patching and attention mechanisms support efficient processing of longer, multivariate histories.

With a patch size of 16, each consecutive block of sixteen weekly observations is embedded into a single token. A 512-week history therefore contributes 32 historical patch tokens per variable, substantially shortening the sequence processed by the transformer. Patching embeds the observations within each block together rather than replacing them with a single temporal average (see model architecture section).

Chronos-2 also separates attention across time from attention across related series. Temporal attention operates across patch positions within each series, whereas group attention operates across series within the same forecasting group at a shared patch position \cite{ansari2025chronos2}. Alternating these operations allows information to propagate across both dimensions without applying unrestricted attention to every pair of variable-time tokens. Together, these architectural choices support the longer context window and expanded multivariate inputs while limiting their computational cost.

\subsection*{Combining observational and simulated training data}

The original Mantis was trained entirely on mechanistically simulated outbreaks. This choice reflected concerns about learning excessively from the diseases and surveillance systems most heavily represented in those data (we had extensive data for respiratory diseases in the United States, specifically), as well as a broader demonstration of proof-of-concept that a mechanistic simulation-based foundation model was feasible. Simulation provided a way to generate a much broader range of epidemic trajectories without repeatedly training on a comparatively narrow observational dataset.

For Mantis 2, we assembled a larger and more diverse collection of observational disease datasets, making it possible to investigate whether real-world training could complement mechanistic simulation. Reservoir \cite{reservoir} provides diverse simulated trajectories, but no simulator can perfectly reproduce the distribution of real surveillance data. Differences may arise from omitted mechanisms, the frequency with which particular dynamics are simulated, or the representation of reporting noise and other observation processes. Observational training offers examples of these characteristics directly from the settings in which the model will be used.

We therefore trained Mantis 2 using both data sources and compared training mixtures ranging from entirely observational to entirely simulated. These experiments showed that combining the two could improve performance over either source alone, while also revealing that the preferred balance depended on the forecasting task and available surveillance history. The final model uses a curriculum that progressively reduces simulated exposure and ends with observational training. The rationale is to expose the model initially to a broad range of mechanistically plausible trajectories and subsequently adapt it to the characteristics of observed surveillance data.

\subsection*{Extending the forecast horizon}

We increased the maximum forecast horizon from eight to sixteen weeks. The motivation was to provide information over a longer planning period for decisions such as staffing, procurement, and hospital capacity preparation. Longer forecasts also allow an anticipated rise, peak, and decline to be represented within a single forecast trajectory, potentially providing information about both the timing and duration of elevated disease activity.

The extended horizon is an additional capability rather than a requirement to use all sixteen forecast weeks. Users can select the portion relevant to their planning needs. Because forecast usefulness depends on lead time and setting, we evaluated performance separately over short, medium, and long horizons.

\subsection*{Joint forecasting of multiple surveillance variables}

The original Mantis forecasts a single target series, optionally informed by one additional covariate. Mantis 2 supports joint forecasting of up to six variables. This change reflects the fact that surveillance systems often provide several related measurements of the same outbreak, such as cases, hospital admissions, emergency department visits, deaths, etc. Their relationships can provide information that is unavailable from any one series alone, including differences in timing between upstream and downstream indicators.

Joint forecasting allows these signals to inform one another within a single model application and produces forecasts for multiple outcomes without requiring a separate run configured around each target. Additional variables can also be supplied as auxiliary inputs when only a subset of outcomes is of interest. The same model remains applicable when only one series is available, allowing its use across settings with different levels of surveillance coverage.

\subsection*{A single model for different forecasting needs}

The original release required users to choose among four separately downloaded models: four- or eight-week forecasting, each with or without a covariate. Although these variants supported different use cases, selecting and managing separate models added complexity to deployment.

Mantis 2 consolidates these capabilities into a single model supporting between one and six variables and forecast horizons from one to sixteen weeks. Users can change the available inputs or requested horizon without downloading a different set of weights. This simplifies integration into forecasting workflows and allows the same deployed model to accommodate changes in surveillance availability and planning needs.

\section{Data}

\subsection*{Synthetic training data}
We generated 20,000 synthetic epidemic trajectories across three transmission mechanisms using a large-scale epidemic simulator called Reservoir \cite{reservoir}. Each simulation spans 5,000 days and contains 6 surveillance modalities: cases, hospitalizations, hospital occupancy, wastewater, syndromic, and deaths. The following epidemiological priors were used to generate the synthetic data set:

\begin{table}[H]
\centering
\begin{tabular}{llll}
\toprule
Quantity & Prior & Notes / clipping & Citation \\
\midrule
Infectious period (d) & $\mathcal{LN}(\log 5, 0.5)$  & clip $[1,30]$ & \cite{hakki2022onset, gani2004epidemiologic} \\
Latent period (d)     & $\mathcal{LN}(\log 5, 0.6)$  & clip $[0.5,21]$ & \cite{kiang2025modeling, xin2022estimating}  \\
Immunity duration (d) & $\mathcal{LN}(\log 365, 1.0)$ & clip $[14, 10950]$ & \cite{edridge2020seasonal, amanna2007duration} \\
$p_a$ (asymptomatic)  & $\text{Beta}(3,7)$           & mean $0.30$& \cite{sah2021asymptomatic, leung2015fraction} \\
$\alpha$ (asymp.\ trans.) & $\text{Beta}(2,5)$       & mean $0.29$& \cite{buitrago2022occurrence} \\
$\kappa$ ($I_a\!\to\!I_s$) & $\exp \mathcal{U}(\log 0.02, \log 0.5)$ && \\
$n_w$ (waning stages) & $\{1,1,1,2,3\}$ uniform pick & mostly $1$ & \cite{wearing2005appropriate} \\
\bottomrule
\end{tabular}
\caption{Shared epidemiological priors (\code{sample\_epi\_params}).}
\label{tab:prior-epi}
\end{table}

\begin{table}[H]
\centering
\begin{tabular}{llll}
\toprule
Mechanism & Quantity & Prior & Cite \\
\midrule
all          & $R_0$ (base)        & $\text{Gamma}(1.2,\,\text{rate}\,1.0)+0.8$ \;(mean $2.0$) & \cite{ke2021estimating, nikbakht2019comparison} \\
             & $R_0$ (vector)      & $\text{Gamma}(3.2,\,\text{rate}\,0.9)+0.4$ \;(mean $3.96$) & \cite{liu2020reviewing} \\
             & $R_0$ (environmental)  & $\text{Gamma}(2.6,\,\text{rate}\,0.85)+0.4$ \;(mean $3.46$) & \cite{mukandavire2011estimating}\\
\midrule
vector       & $\sigma_v$          & $\exp\mathcal{U}(\log\tfrac1{21}, \log\tfrac14)$ & \cite{liu2020reviewing} \\
             & $\mu_v$             & $\exp\mathcal{U}(\log\tfrac1{30}, \log\tfrac17)$  & \cite{liu2020reviewing} \\
             & $a$ (biting)    & $\mathcal{U}(0.1,1.0)$  &  \cite{liu2020reviewing}\\
             & $b_h, b_v$         & $\mathcal{U}(0.1,0.8)$ each &  \cite{liu2020reviewing} \\
             & $f_{Nv}, f_{Iv}$    &  $\mathcal{U}(0.05,2.0),\;\mathcal{U}(0.001,0.05)$ & \cite{liu2020reviewing}\\
\midrule
environmental   & environmental share    & $\mathcal{U}(0.1,0.9)$ of $R_0$ & \cite{mukandavire2011estimating} \\
             & $\eta_I$            & $\exp\mathcal{U}(\log 0.1, \log 1.0)$  & \cite{mukandavire2011estimating} \\
             & $\eta_A$            & $\exp\mathcal{U}(\log 0.05, \log 0.5)$ & \cite{mukandavire2011estimating} \\
             & $\alpha_\text{env}$ & $\mathcal{U}(0.1,1.0)$ & \cite{mukandavire2011estimating}  \\
             & $\mu_w$             & $\exp\mathcal{U}(\log 0.1, \log 0.5)$ & \cite{mukandavire2011estimating}  \\
\bottomrule
\end{tabular}
\caption{Mechanism-specific priors. All $R_0$ draws are floored at $0.3$. Transmission constants $\beta$ and $\delta$ are back-calculated from $R_0$.}
\label{tab:prior-mech}
\end{table}

\subsection*{Real-world training data}
We assembled a multi-source epidemiological surveillance dataset spanning eight pathogen surveillance systems across four continents. Thai provincial surveillance data (2003--2019) provided weekly case counts for 20 notifiable diseases across 77 provinces, including enteric, vectorborne, respiratory, and bloodborne pathogens. Historical US disease surveillance was sourced from Project Tycho Level 2 \cite{project_tycho}, yielding 39 diseases with state- and regional-level weekly case counts spanning 1888--2014, with disease-specific active windows defined by years exceeding 5\% of peak annual incidence. International dengue surveillance was drawn from the Open Dengue dataset \cite{clarke2024global}, comprising 59 country-level series and two sub-national datasets (Nicaragua, $L=18$ provinces; Thailand, $L=78$ provinces), with monthly data interpolated to weekly resolution via smoothing spline. US notifiable disease surveillance (2022-2026) was obtained from the National Notifiable Diseases Surveillance System (NNDSS), contributing 54 disease-location combinations at regional and national levels. Influenza-like illness (ILI) and related respiratory surveillance data for 148 countries were obtained from WHO FluNet (1996--2026), providing up to five simultaneous signals per country (ILI, SARI, ARI, pneumonia cases, SARI deaths). Mpox case and death counts were obtained from WHO situation reports (2022--2026) for 33 countries and six WHO regional aggregates. Ebola virus disease data for the 2018--2020 DRC outbreak were sourced from the Humanitarian Data Exchange, aggregated from daily health-zone reports to weekly provincial counts. Cholera surveillance data were obtained from two sources: WHO AFRO weekly reports (2023--2026, 10 countries, cases and deaths) and a JHU-compiled dataset (2010--2023, 32 countries, cases only). US influenza and RSV multivariate surveillance series were constructed by joining five Delphi Epidata signals (wILI, percent positivity, hospitalizations, deaths, and ED visit percentage) at the state level (2010--2026).

All datasets were standardized to a common blinded format: a two-dimensional array of shape $(T \times L,\; 9)$, where $T$ denotes the number of weekly timesteps, $L$ the number of spatial locations, and the 9 columns encode $(\text{scenario\_index},\; \sin\omega_t,\; \cos\omega_t,\; v_1, \ldots, v_6)$. The cyclical week-of-year encoding is defined as $\sin(2\pi w / 52.18)$ and $\cos(2\pi w / 52.18)$, where $w$ is the ISO week number. The primary signal ($v_1$) is always case counts; secondary signals ($v_2$--$v_6$) encode deaths or additional surveillance streams where available, and $-1$ otherwise. Explicit reported zeros are preserved as $0$; gaps of four weeks or fewer within a reporting window are filled by cubic spline interpolation; gaps exceeding four weeks and periods outside a location's active reporting window are assigned the value $-1$.

\subsection*{Evaluation Data}
The evaluation set comprises two components. The in-domain evaluation set consists of 24 disease-location combinations drawn from the real-world training set and held out temporally (208 week evaluation window). The out-of-domain evaluation set comprises 12 disease-location combinations that include  COVID-19, two healthcare-associated infection series from NNDSS, dengue in Ecuador, meningitis and smallpox from Project Tycho, influenza ILI in Austria, measles in Thailand, mpox in Sierra Leone, ebola in the DRC, and cholera in Ethiopia. Out-of-domain series test zero-shot generalization to novel pathogens, geographies, and surveillance modalities.

\section{Model architecture}

\subsection*{Chronos-2 backbone}

Mantis 2 uses the Chronos-2 transformer backbone \cite{ansari2025chronos2}, initialized from the publicly released weights. The model contains twelve transformer layers with a hidden dimension of 768 and approximately 121 million parameters, including the latent feedback module described below. We train the full model on infectious disease trajectories.

Chronos-2 combines temporal attention within individual series with group attention across related series at corresponding patch positions. In Mantis 2, each forecasting task forms a group containing the selected disease variables and two calendar covariates. Attention shares information within each group, while separate forecasting tasks in the same batch remain isolated (i.e., each group is a multivariate forecasting task).

\subsection*{Input representation and forecast outputs}

The model receives up to 512 weeks of historical observations for between one and six disease variables. Two additional series encode the annual calendar cycle using sine and cosine covariates. These covariates are supplied over both the historical context and the sixteen-week forecast horizon, allowing the model to condition on the seasonal position of future observations. Future disease observations are not supplied as inputs.

Disease variables are transformed using the inverse hyperbolic sine function, before being passed to the model's instance-normalization. This transformation compresses large values while remaining approximately linear near zero. Calendar covariates are supplied without this external transformation.

Historical inputs are divided into non-overlapping patches of sixteen weeks. A full 512-week context therefore occupies 32 historical patch positions per series. Each patch embedding represents its values together with  temporal information. The sixteen-week forecast is represented by one future patch per series. Disease-target values in this patch are unknown, whereas the calendar covariate values are available. The output head predicts 21 quantiles at each of the sixteen forecast weeks directly in one shot (i.e., without autoregressively feeding predicted weekly values back into the model). The model then reverses its internal normalization, and the external transformation is inverted using the hyperbolic sine function to return forecasts in the original units.

\subsection*{Latent feedback recurrence}

We augment the backbone with a latent feedback mechanism that allows additional processing before the forecast is produced \cite{latentcot}. The mechanism operates on hidden representations associated with the future patch. It does not generate textual reasoning or intermediate forecasts, and no intermediate reasoning targets are supplied during training (i.e., we don't tell the model what to do with the intermediate tokens, we only supervise the final prediction).

The model first processes the historical patches and the future patch through the encoder. For each series, the resulting hidden representation at the future-patch position is passed through a feedback MLP. Both the input and hidden widths are 768. The resulting vector is multiplied by a learned scalar gate parameterized through a hyperbolic tangent. The effective gate value is initialized to 0.95.

The gated vector is appended as an additional token after the original input sequence, and the same encoder is run again. A second feedback token is then constructed from the updated representation at the original future-patch position. Both feedback tokens are retained for a third encoder pass. Thus, the final configuration uses two feedback recurrences and three encoder passes in total, with shared encoder and feedback-module weights across passes.

Each recurrent pass receives the original input embeddings together with the accumulated feedback tokens. Feedback tokens are available to the backbone's temporal and group attention mechanisms. After the final pass, the output head reads the updated representation at the original future-patch position; the feedback tokens themselves are not decoded into forecasts. With a full context, the sequence grows from 34 positions in the initial pass to 36 positions in the final pass: 32 historical patches, one register token, one future patch, and two feedback tokens per series.

Training gradients propagate through all three encoder passes and the feedback construction steps. The feedback module and gate are learned jointly with the pretrained backbone using the final forecasting objective, without a separate loss for intermediate representations.

\section{Model training}

\subsection*{Training-example construction}

Training uses separate pools of simulated and observational trajectories prepared ahead of time. Each trajectory contains weekly observations for up to six disease variables together with the two calendar covariates.

A training task is constructed by first selecting the data source (synthetic or real) according to the training schedule and then sampling a trajectory uniformly from that source. For a trajectory of length $T$, a forecast origin $t$ is sampled uniformly from the eligible positions satisfying $8 \leq t \leq T-16$. The historical input contains the observations preceding $t$, truncated to the most recent 512 weeks. Shorter histories are right-aligned and padded on the left with missing values. The target consists of the next sixteen weekly observations.

For each sampled origin, we identify disease variables with at least one observed value in the context window. We sample the number of included variables uniformly between one and the number available, then select that many variables without replacement. All selected variables are forecast targets during training. This procedure exposes the model to different combinations of surveillance signals and to both univariate and multivariate tasks. The two calendar covariates are included in every task, with their future values supplied as known inputs.

Forecast origins and variable subsets are resampled throughout training. Consequently, training repeatedly draws different tasks from the available trajectories rather than making a single ordered pass through a fixed collection of forecasting windows.

\subsection*{Simulation-to-observation curriculum}

The schedule begins with 70\% simulated training and ends with entirely observational training. Its purpose is to expose the model initially to a broad distribution of mechanistically plausible epidemic trajectories and progressively emphasize the statistical characteristics of real surveillance data. Averaged over the training schedule, the expected simulated fraction is 35\% of sampled tasks.

\subsection*{Forecasting objective and optimization}

The model is trained using the quantile loss (also called pinball loss) which averages over all quantiles $\tau$:

\begin{equation*}
L_\tau(y, \hat{y}) = \begin{cases} 
\tau (y - \hat{y}) & \text{if } y \ge \hat{y} \\ 
(\tau - 1)(y - \hat{y}) & \text{if } y < \hat{y} 
\end{cases}
\end{equation*}

The backbone applies its loss computation to normalized targets and predictions across forecast weeks and quantile levels. The latent feedback mechanism is optimized through this same final forecast loss.

Optimization uses AdamW \cite{adamw} with a peak learning rate of $3\times10^{-6}$ and weight decay of 0.01. The learning rate increases during the first 500 optimizer steps and then follows a cosine decay over the remainder of the 8,000-step run. Gradients are clipped to a maximum global norm of 1.0 before each optimizer update. The model is trained on a single NVIDIA RTX PRO 6000 Blackwell Server Edition GPU.

\section{Model evaluation}

\subsection*{Evaluation datasets and forecast origins}

We evaluated forecasts on disease-location datasets containing one or more surveillance trajectories. For datasets represented during Mantis 2 training, the final 4 years of each series was reserved for evaluation. For out-of-domain evaluation, entire disease-location datasets were excluded from Mantis 2 training. Data sources and preprocessing are described in the data section. Evaluation used rolling forecast origins separated by four weeks. Forecast windows therefore overlap.

All received up to the most recent 512 weeks of context. Shorter histories were right-aligned and padded, with missingness represented through the model-specific input procedure. A variable was eligible for evaluation at an origin only if it contained at least one finite observation within this context window.

Designated surveillance variables, such as cases or deaths, were scored as forecast targets. Columns that contained useful data that was not infectious disease outcomes (e.g., weather) were supplied as historical inputs to the multivariate models but excluded from scoring. Calendar sine and cosine covariates were supplied over both the historical context and forecast horizon. Future surveillance observations were used only for scoring.

\subsection*{Pretrained forecasting models}

Mantis 2 was evaluated using its final training checkpoint, without any task-specific parameter updates. Chronos-2 was evaluated using the publicly released pretrained weights, also without task-specific fitting.

\subsection*{Persistence baseline}

One baseline we evaluated was a naive persistence baseline \cite{cramer2022covid}. The persistence baseline used the most recent observed target value as its median forecast at every lead time. Uncertainty was generated using a symmetric random walk based on historical changes, following the construction of the COVIDhub-baseline \cite{cramer2022covid}.

\subsection*{Seasonal-naive baseline}

The seasonal-naive baseline tests whether a forecasting model improves on simply repeating the previous year's disease activity. This provides a useful reference for diseases with recurring seasonal patterns: a model may anticipate seasonal rises and declines without learning anything beyond their annual recurrence.

For each forecast week, the baseline uses the observation from 52 weeks earlier as its central prediction. Forecast uncertainty is estimated from historical differences between observations separated by 52 weeks. Larger year-to-year differences produce wider prediction intervals, reflecting how reliably the previous year's trajectory predicts subsequent activity.

When the corresponding observation from the previous year is unavailable, the baseline cannot generate a forecast for that week. We exclude the seasonal-naive baseline from an origin--horizon comparison if it cannot forecast every target value included in that comparison, rather than substituting a different forecasting method.

\subsection*{Exponential smoothing (ETS)}

Exponential smoothing provides a statistical baseline that forecasts from estimates of the current disease activity level, its trend, and, when sufficient history is available, its seasonal pattern. These estimates are updated by weighting recent observations more heavily than older observations. We fitted a separate model to each trajectory and scored target using all target history preceding the fitting origin in $\log(1+y)$ space.

We used a hierarchy of model configurations to accommodate differences in the amount of available history and to provide forecasts when more complex models could not be fitted. With at least 104 observations, corresponding to two annual cycles, we first fitted a model with additive seasonality of period 52 and a damped additive trend. The seasonal component captures recurring annual variation, while damping progressively reduces the contribution of the current trend at longer forecast horizons.

When fewer than two annual cycles were available, or fitting the seasonal model raised an error, we used a nonseasonal model with a damped additive trend, provided at least ten observations were available. If this configuration was unavailable or fitting raised an error, we tried an undamped additive trend with at least four observations, followed by simple exponential smoothing, which models the activity level without an explicit trend or seasonal component. We used the first configuration that fitted without an error. This hierarchy provided a consistent fitting procedure across datasets.

To construct probabilistic forecasts, we estimated a residual scale from the root mean squared in-sample residuals and placed Gaussian uncertainty intervals around the fitted forecast trajectory. The uncertainty scale increased in proportion to the square root of the lead time from the fitting origin. Forecast quantiles were floored at zero in transformed space.

\subsection*{LSTM model}

The LSTM baseline was trained separately for each disease-location dataset, pooling training windows across trajectories within that dataset. Its recurrent component contained three LSTM layers with 64 hidden units per layer and dropout of 0.1 between layers.

At each historical time point, the input contained six variable values, two calendar covariates, and six binary indicators of whether the variable values were observed, giving fourteen input features. Missing values and left-padding were represented by zeros, with the observation indicators distinguishing missing target values from observed zeros. Inputs were represented in $\log(1+y)$ space.

The final hidden state of the top LSTM layer was concatenated with the calendar covariates for each forecast week. A shared prediction head, consisting of a 64-unit linear layer, a ReLU activation, and an output layer, produced quantiles for all six variable slots at each lead. Only designated surveillance targets contributed to the training loss.

At each refit, the model was initialized from scratch and trained on pooled historical windows. Candidate windows were ordered by their forecast-origin index, and at most the 2,048 most recent candidates were retained.

Training minimized quantile loss using AdamW with learning rate $10^{-3}$ and minibatches of 32 windows. Each fit ran for up to 50 epochs, subject to a maximum of 2,000 optimizer updates, with gradient norms clipped at 1.0 (these were necessary to prevent overfitting). At inference, quantiles were sorted to remove crossings and floored at zero.

\subsection*{Seasonal autoregressive integrated moving average (SARIMA)}

Seasonal autoregressive integrated moving average (SARIMA) models provide a statistical baseline that captures relationships between current disease activity, previous observations, and previous prediction errors. Differencing allows the model to represent changes in activity rather than assuming a constant long-term level, while seasonal terms capture dependence between observations separated by an annual cycle.

We fitted a separate univariate model to each trajectory and scored target. Models used the target's own history without additional surveillance variables or exogenous covariates. Similar to exponential smoothing, fitting was performed in $\log(1+y)$ space, leading missing values were removed, and remaining gaps were filled within the available fitting history.

With at least 104 observations, corresponding to two annual cycles, we first fitted
\begin{equation*}
\operatorname{SARIMA}(1,1,1)(1,0,0)_{52}.
\end{equation*}
This specification includes one nonseasonal autoregressive term, one order of differencing, one nonseasonal moving-average term, and one seasonal autoregressive term at lag 52. It therefore captures both short-term dependence and annual recurrence, without seasonal differencing or a seasonal moving-average term.

When fewer than 104 observations were available, or the seasonal model could not produce a valid fit and forecast, we used a nonseasonal $\operatorname{ARIMA}(1,1,1)$ model. If that model also failed, we used the simpler $\operatorname{ARIMA}(0,1,1)$ specification. This fixed hierarchy allowed the baseline to accommodate short histories and fitting difficulties without conducting a separate order search for every trajectory and forecast origin. Models were fitted without a deterministic trend. Stationarity constraints were imposed on the autoregressive components and invertibility constraints on the moving-average components.

\subsection*{Refitting schedules and information updates}

Forecasts were evaluated every four weeks, but the locally fitted models were refitted every ten forecast horizons to limit the already high computational cost of repeated fitting. The LSTM, ETS, and SARIMA models therefore used fitting origins at weeks 8, 48, 88, and so on. At each fitting origin, the models were fitted using the preceding observations.

The models differed in how they incorporated observations arriving between refits. The LSTM retained its fitted parameters but received updated historical context at every evaluation origin. Its forecasts therefore reflected the latest observations even when its parameters had not recently been retrained.

Persistence and seasonal-naive forecasts were recomputed from the available history at every evaluation origin. Mantis 2 and Chronos-2 likewise received updated context at each origin, while their pretrained parameters remained fixed throughout evaluation.

\subsection*{Forecast scores and matched comparisons}

We evaluated forecast accuracy over three nested horizon ranges: weeks 1-4, 1-8, and 1-16. These ranges measure cumulative performance from the first forecast week through each specified endpoint. We used weighted interval score (WIS) \cite{wis} to assess probabilistic forecasts. All models used the same quantile grid for forecasting.

Scores were calculated in $\log(1+y)$ space, which reduces the influence of large absolute differences in disease activity across surveillance targets. The reported scores measure accuracy on this transformed scale rather than in original case, hospitalization, or death counts. At each forecast origin and horizon range, we identified the finite target observations across the eligible surveillance variables. These observations defined a common scoring set for all models. A model received a score for an origin and horizon range only if it provided finite predictions for every required quantile at every target value in the common scoring set. This prevented a model from receiving a score based only on a subset of the targets evaluated for other models. Whenever two models had results for the same dataset, trajectory, origin, and horizon range, their scores therefore referred to exactly the same target values.

\section{Full forecasting results}
\label{appendix:full_results}

We report aggregate weighted interval score (WIS) across three cumulative forecast horizon ranges in Table \ref{tab:overall_wis}, empirical prediction interval coverage in Table \ref{tab:overall_coverage}, and WIS for each evaluation dataset over weeks 1-16 in Table \ref{tab:dataset_wis}. WIS was
calculated in $\log(1+y)$ space, with lower values indicating better probabilistic forecast performance. Aggregate scores weight forecasted target values equally, following the procedure described in the model-evaluation section.

\subsection*{Aggregate probabilistic forecast performance}

Mantis 2 achieved the lowest reported aggregate WIS in all three horizon ranges, followed by Chronos-2 and persistence (Table \ref{tab:overall_wis}). Mantis 2's aggregate WIS was 0.20913 over weeks 1-4, 0.28439 over weeks 1-8, and 0.36965 over weeks 1-16. The corresponding Chronos-2 scores were 0.23097, 0.31084, and 0.39762. Scores increased over longer horizon ranges for every model, reflecting the greater difficulty of forecasting further into the future.

These summaries include all available forecasts for each model. Forecast availability differed among models, so the aggregate values do not all represent identical evaluation samples. In particular, seasonal-naive forecasts require observations from the corresponding weeks of the preceding year and are unavailable when those observations are absent. These tables provide descriptive summaries of available forecasts; comparisons restricted to common forecast instances follow the matching procedure described in the evaluation methods.

\begin{table}[htbp]
\centering
\caption{
Aggregate weighted interval score (WIS) over three cumulative forecast horizon ranges. Scores are calculated in $\log(1+y)$ space and pooled across available forecasted target values. Lower values indicate better performance. Each model's summary includes its available forecasts, which need not constitute the same evaluation sample across models.
}
\label{tab:overall_wis}
\begin{tabular}{lrrr}
\toprule
Model & Weeks 1--4 & Weeks 1--8 & Weeks 1--16 \\
\midrule
Mantis 2            & 0.20913 & 0.28439 & 0.36965 \\
Chronos-2           & 0.23097 & 0.31084 & 0.39762 \\
Persistence         & 0.28046 & 0.36928 & 0.46776 \\
LSTM                & 0.41975 & 0.45859 & 0.51038 \\
Seasonal naive      & 0.52003 & 0.52281 & 0.52770 \\
SARIMA              & 0.59501 & 0.62158 & 0.65710 \\
ETS                 & 0.72205 & 0.75606 & 0.80579 \\
\bottomrule
\end{tabular}
\end{table}

\subsection*{Prediction interval coverage}

Mantis 2's empirical 95\% prediction interval coverage remained close to nominal across the three horizon ranges: 96.14\% over weeks 1--4, 95.11\% over weeks 1--8, and 94.16\% over weeks 1--16 (Table \ref{tab:overall_coverage}). Its central 50\% prediction intervals showed some undercoverage, however, with empirical coverage decreasing from 48.59\% over weeks 1-4 to 45.68\% over weeks 1-16. Thus, agreement with nominal coverage for the wider intervals did not extend equally to the central portion of the predictive distribution.

Chronos-2 also showed declining 50\% coverage over longer horizon ranges. Persistence and seasonal naive achieved approximately correct 50\% coverage but undercovered at the 95\% level. SARIMA exhibited the opposite pattern: its 95\% coverage was close to nominal, while its 50\% coverage was substantially above nominal. ETS and LSTM undercovered at both levels across all three horizon ranges.

Coverage describes how often prediction intervals contain the observations, but does not account for interval width. Consequently, coverage should be interpreted alongside WIS, which evaluates both interval width and penalties for observations falling outside the intervals.

\begingroup
\small
\setlength{\tabcolsep}{6pt}
\begin{longtable}{llrrrr}
\caption{
Empirical coverage of central 50\% and 95\% prediction intervals, expressed as percentages. Values closer to 50\% and 95\%, respectively, indicate closer agreement with nominal coverage. $N$ is the number of forecasted target values scored, and datasets denotes the number of evaluation files contributing results. Values forecast from different origins contribute separately to $N$. Results use all available forecasts for each model; differences in $N$ indicate differences in evaluation availability.
}
\label{tab:overall_coverage} \\
\toprule
Horizon & Model & 50\% coverage & 95\% coverage
& $N$ & Datasets \\
\midrule
\endfirsthead

\multicolumn{6}{l}{
\tablename~\thetable{} --- continued
} \\
\toprule
Horizon & Model & 50\% coverage & 95\% coverage
& $N$ & Datasets \\
\midrule
\endhead

\midrule
\multicolumn{6}{r}{Continued on next page} \\
\endfoot

\bottomrule
\endlastfoot

Weeks 1--4
& Mantis 2       & 48.59 & 96.14 & 164,095 & 30 \\
& Chronos-2      & 49.14 & 96.16 & 164,095 & 30 \\
& Persistence    & 50.72 & 90.77 & 164,095 & 30 \\
& LSTM           & 40.59 & 85.54 & 164,095 & 30 \\
& Seasonal naive & 52.63 & 90.73 & 133,006 & 29 \\
& SARIMA         & 64.60 & 93.94 & 158,219 & 30 \\
& ETS            & 45.63 & 73.66 & 158,219 & 30 \\
\midrule

Weeks 1--8
& Mantis 2       & 46.65 & 95.11 & 328,184 & 30 \\
& Chronos-2      & 46.60 & 94.60 & 328,184 & 30 \\
& Persistence    & 50.62 & 88.72 & 328,184 & 30 \\
& LSTM           & 38.84 & 83.57 & 328,184 & 30 \\
& Seasonal naive & 52.64 & 90.37 & 265,781 & 29 \\
& SARIMA         & 65.45 & 94.25 & 316,432 & 30 \\
& ETS            & 45.82 & 73.62 & 316,432 & 30 \\
\midrule

Weeks 1--16
& Mantis 2       & 45.68 & 94.16 & 656,236 & 30 \\
& Chronos-2      & 45.60 & 93.78 & 656,236 & 30 \\
& Persistence    & 52.86 & 88.44 & 656,236 & 30 \\
& LSTM           & 36.90 & 81.19 & 656,236 & 30 \\
& Seasonal naive & 52.50 & 89.92 & 530,744 & 29 \\
& SARIMA         & 67.65 & 94.87 & 632,845 & 30 \\
& ETS            & 46.94 & 74.41 & 632,845 & 30 \\

\end{longtable}
\endgroup

\subsection*{Performance across individual datasets}

The dataset-level results show variation in the relative performance of the forecasting methods (Table \ref{tab:dataset_wis}). Mantis 2 had lower reported WIS than Chronos-2 on 26 of the 30 datasets. The four exceptions were Rocky Mountain spotted fever, hepatitis B from Project Tycho, salmonellosis, and hepatitis A across U.S. states. Improvements over Chronos-2 were particularly pronounced for carbapenem-resistant Enterobacteriaceae and meningitis.

Persistence had lower reported WIS than both foundation models on COVID-19 and dengue in Nicaragua and Singapore. These results illustrate that the aggregate advantage of Mantis 2 does not imply an advantage on every individual dataset. However, for COVID-19, there were data quality issues (changing definitions for variables, high amounts of missing data) that substantially harmed the performance of the models).

\begingroup
\small
\setlength{\tabcolsep}{3pt}
\renewcommand{\arraystretch}{1.12}
\begin{longtable}{
@{}
>{\raggedright\arraybackslash}p{4.0cm}
rrrrrrr
@{}
}
\caption{
Weighted interval score (WIS) for each evaluation dataset over weeks 1--16. Scores are calculated in $\log(1+y)$ space and pooled over available forecasted target values within each dataset; lower values indicate better performance. Seasonal denotes the seasonal-naive baseline. Model-specific forecast availability can differ within a dataset, so the entries are summaries of available forecasts rather than necessarily matched comparisons.
}
\label{tab:dataset_wis} \\
\toprule
Dataset & Mantis 2 & Chronos-2 & Persistence
& LSTM & Seasonal & SARIMA & ETS \\
\midrule
\endfirsthead

\multicolumn{8}{l}{
\tablename~\thetable{} --- continued
} \\
\toprule
Dataset & Mantis 2 & Chronos-2 & Persistence
& LSTM & Seasonal & SARIMA & ETS \\
\midrule
\endhead

\midrule
\multicolumn{8}{r}{Continued on next page} \\
\endfoot

\bottomrule
\endlastfoot

COVID-19
& 0.70492 & 0.70515 & 0.65241 & 1.11491
& 0.76370 & 0.73894 & 0.96699 \\

Dengue--Ecuador
& 0.28054 & 0.28172 & 0.39999 & 0.54665
& 0.51285 & 0.40595 & 0.41814 \\

Dengue--Indonesia
& 0.21717 & 0.22298 & 0.22676 & 0.46587
& 0.45524 & 0.18264 & 0.20858 \\

Dengue--Nicaragua
& 0.41206 & 0.42872 & 0.40332 & 0.56666
& 0.65005 & 0.41656 & 0.83064 \\

Dengue--Singapore
& 0.30825 & 0.31734 & 0.29664 & 0.56918
& 0.81328 & 0.31197 & 0.68681 \\

Dengue--Thailand
& 0.44336 & 0.52093 & 0.58610 & 0.61893
& 0.87349 & 0.57618 & 0.80000 \\

Influenza--Austria
& 0.53134 & 0.57315 & 0.83065 & 0.79190
& 0.78175 & 0.77091 & 0.91781 \\

Influenza--Nigeria
& 0.39054 & 0.41663 & 0.51245 & 0.41993
& 0.54459 & 0.39882 & 0.87472 \\

C. auris
& 0.35850 & 0.42594 & 0.58217 & 0.62885
& 0.60519 & 0.41175 & 0.56139 \\

Carbapenem-resistant Enterobacteriaceae
& 0.38580 & 0.72789 & 0.70474 & 0.72491
& 0.71342 & 0.56367 & 1.22310 \\

Acute diarrhea
& 0.10257 & 0.11815 & 0.16467 & 0.18041
& 0.18429 & 0.21636 & 0.23275 \\

Hand, foot and mouth disease
& 0.37564 & 0.41699 & 0.51941 & 0.50039
& 0.62843 & 0.49484 & 0.57119 \\

Hepatitis B--Thailand
& 0.24038 & 0.26376 & 0.30042 & 0.29091
& 0.36324 & 0.32481 & 0.47011 \\

Influenza--Thailand
& 0.33909 & 0.38604 & 0.49060 & 0.49255
& 0.57972 & 0.48959 & 0.59103 \\

Tuberculosis--Thailand
& 0.23959 & 0.27683 & 0.32609 & 0.32507
& 0.47704 & 0.34416 & 0.46552 \\

Enteric fever--Thailand
& 0.21153 & 0.23184 & 0.28508 & 0.36154
& 0.32424 & 0.33351 & 0.40330 \\

Hepatitis A--Thailand
& 0.23734 & 0.24533 & 0.26784 & 0.31329
& 0.34513 & 0.28079 & 0.41864 \\

Measles--Thailand
& 0.20972 & 0.23775 & 0.26850 & 0.27940
& 0.45435 & 0.29050 & 0.39180 \\

Rocky Mountain spotted fever
& 0.49137 & 0.48797 & 0.68073 & 0.51322
& 0.55134 & 0.56363 & 0.65229 \\

Brucellosis
& 0.26475 & 0.28173 & 0.39602 & 0.32074
& 0.33971 & 0.28188 & 0.38612 \\

Hepatitis B--U.S. regions
& 0.16503 & 0.15991 & 0.31458 & 0.27840
& 0.24020 & 0.16469 & 0.29659 \\

Malaria
& 0.31723 & 0.32845 & 0.46807 & 0.33991
& 0.38766 & 0.33315 & 0.54008 \\

Meningitis
& 0.37116 & 0.52148 & 0.56737 & 0.42915
& 0.47907 & 0.44849 & 1.11943 \\

Salmonellosis
& 0.24097 & 0.23614 & 0.37260 & 0.38237
& 0.25999 & 0.33638 & 0.45154 \\

Smallpox
& 0.32266 & 0.34331 & 0.54725 & 0.46356
& 0.48240 & 0.46900 & 0.72112 \\

Typhoid fever
& 0.28638 & 0.29434 & 0.48861 & 0.34233
& 0.35606 & 0.33680 & 0.52707 \\

Pertussis
& 0.38346 & 0.39172 & 0.52775 & 0.45483
& 0.61361 & 0.39053 & 0.53248 \\

Hepatitis A--U.S. states
& 0.37582 & 0.37016 & 0.54197 & 0.39382
& 0.49754 & 0.39038 & 0.60722 \\

Measles--U.S. states
& 0.45860 & 0.49354 & 0.59108 & 0.50994
& 0.66532 & 0.56298 & 0.76852 \\

Tuberculosis--U.S. states
& 0.36580 & 0.38778 & 0.63001 & 0.39505
& 0.50698 & 0.42209 & 0.76121 \\

\end{longtable}
\endgroup

\section{Comparison with the original Mantis}
\label{appendix:mantis_comparison}

Mantis 2 expands the forecasting capabilities of the original Mantis, but an important additional question is whether it improves accuracy on tasks its predecessor already supports. The original Mantis forecasts a single outcome up to eight weeks ahead, optionally using one additional covariate \cite{mantis}. It therefore cannot participate directly in the full sixteen-week, joint multivariate evaluation used in the main analysis. We report a separate comparison restricted to single-target forecasting over eight weeks, allowing us to assess improvements within the original model's forecasting scope.

\subsection*{Comparison design}

We included evaluation datasets containing exactly one populated surveillance target. This restriction yielded 27 eligible datasets: the COVID-19 dataset and the influenza datasets from Austria and Nigeria were excluded because they contained multiple surveillance targets. Auxiliary variables could still be supplied as inputs, but only the designated surveillance target was scored.

Both models were evaluated at the same forecast origins, beginning after eight weeks of observations and advancing in four-week increments. We retained the main evaluation's requirement that sixteen subsequent weeks be available, preserving its forecast origins, but scored only weeks one through eight. Original Mantis generated an eight-week forecast directly; Mantis 2 generated its usual sixteen-week forecast, from which the first eight weeks were extracted.

Original Mantis used its released eight-week checkpoints. At each origin, the covariate-aware checkpoint was used when an auxiliary variable was available within its historical context; otherwise, the covariate-free checkpoint was used. When several auxiliary variables were available, the first eligible variable in the dataset's column order was selected, without selection based on forecast performance. Original Mantis received up to 112 weeks of history.

Mantis 2 received up to 512 weeks of history, all available auxiliary variables, and its usual calendar covariates. Neither model underwent task-specific retraining. Thus, the comparison matched forecast targets, origins, and horizons while retaining each model's supported input configuration. It evaluates the overall improvement offered by Mantis 2, including its ability to use additional historical and auxiliary information, rather than isolating an architectural change under identical inputs.

Percentage improvement was defined as
\begin{equation*}
100 \times
\frac{
\mathrm{WIS}_{\mathrm{Mantis}}
-
\mathrm{WIS}_{\mathrm{Mantis\,2}}
}{
\mathrm{WIS}_{\mathrm{Mantis}}
}.
\end{equation*}
Positive values indicate lower forecast error for Mantis 2.

\subsection*{Results}

Across 34,762 matched forecast origins, Mantis 2 reduced pooled WIS from 0.351921 to 0.241338, an improvement of 31.42\%. Both models returned valid forecasts for every evaluated pair. Mantis 2 achieved lower WIS on all 27 eligible datasets, with dataset-level improvements ranging from 10.37\% to 45.55\% (Table \ref{tab:mantis_comparison}).

Improvements extended across diseases, locations, and surveillance settings. WIS decreased by 45.55\% for historical U.S. brucellosis, 44.66\% for acute diarrhea in Thailand, and 43.27\% for measles across U.S. states. All five dengue datasets improved, with reductions ranging from 10.55\% in Nicaragua to 39.91\% in Indonesia. Gains occurred both in datasets without auxiliary variables and in those where original Mantis used its covariate-aware checkpoint.

These results show that Mantis 2's advances extend beyond supporting longer horizons and jointly forecasting multiple outcomes: it also substantially improves probabilistic accuracy within the eight-week, single-target setting supported by its predecessor. Because the models differ in training data, architecture, historical context, and auxiliary inputs, this comparison does not attribute the improvement to any one component.

\begingroup
\small
\setlength{\tabcolsep}{5pt}
\renewcommand{\arraystretch}{1.1}
\begin{longtable}{@{}p{7.0cm}rrr@{}}
\caption{
Comparison with the original Mantis over forecast weeks 1--8.
WIS is calculated in $\log(1+y)$ space using nine common
quantiles and matched forecast targets and origins.
Lower WIS is better. Improvement is the percentage reduction
in WIS relative to original Mantis. Models retain their
respective context lengths and auxiliary-input configurations.
}
\label{tab:mantis_comparison} \\
\toprule
Dataset & Mantis & Mantis 2 & Improvement (\%) \\
\midrule
\endfirsthead

\multicolumn{4}{l}{\tablename~\thetable{} --- continued} \\
\toprule
Dataset & Mantis & Mantis 2 & Improvement (\%) \\
\midrule
\endhead

\midrule
\multicolumn{4}{r}{Continued on next page} \\
\endfoot

\bottomrule
\endlastfoot

Dengue---Ecuador
& 0.24591 & 0.19890 & 19.12 \\
Dengue---Indonesia
& 0.17306 & 0.10400 & 39.91 \\
Dengue---Nicaragua
& 0.34377 & 0.30749 & 10.55 \\
Dengue---Singapore
& 0.24938 & 0.20207 & 18.97 \\
Dengue---Thailand
& 0.42178 & 0.28882 & 31.52 \\
\textit{Candida auris}---U.S.
& 0.35623 & 0.31928 & 10.37 \\
Carbapenem-resistant Enterobacteriaceae---U.S.
& 0.58237 & 0.36914 & 36.61 \\
Acute diarrhea---Thailand
& 0.12314 & 0.06815 & 44.66 \\
Hand, foot, and mouth disease---Thailand
& 0.34725 & 0.26642 & 23.28 \\
Hepatitis B---Thailand
& 0.28187 & 0.19936 & 29.27 \\
Influenza---Thailand
& 0.32083 & 0.23502 & 26.75 \\
Tuberculosis---Thailand
& 0.26999 & 0.17999 & 33.33 \\
Enteric fever---Thailand
& 0.22731 & 0.16108 & 29.14 \\
Hepatitis A---Thailand
& 0.24302 & 0.20300 & 16.47 \\
Measles---Thailand
& 0.21477 & 0.13684 & 36.28 \\
Rocky Mountain spotted fever---U.S.
& 0.49390 & 0.42352 & 14.25 \\
Brucellosis---U.S.
& 0.46422 & 0.25276 & 45.55 \\
Hepatitis B---U.S.
& 0.20465 & 0.15700 & 23.28 \\
Malaria---U.S.
& 0.42480 & 0.29652 & 30.20 \\
Meningitis---U.S.
& 0.39900 & 0.32117 & 19.51 \\
Salmonellosis---U.S.
& 0.24642 & 0.21008 & 14.75 \\
Smallpox---U.S.
& 0.37870 & 0.28199 & 25.54 \\
Typhoid fever---U.S.
& 0.33103 & 0.26223 & 20.78 \\
Pertussis---U.S.
& 0.38222 & 0.33128 & 13.33 \\
Hepatitis A---U.S. states
& 0.51480 & 0.34810 & 32.38 \\
Measles---U.S. states
& 0.69481 & 0.39416 & 43.27 \\
Tuberculosis---U.S. states
& 0.51103 & 0.34793 & 31.92 \\
\midrule
\textbf{Pooled}
& \textbf{0.35192} & \textbf{0.24134} & \textbf{31.42} \\

\end{longtable}
\endgroup

\section{Training-mixture and surveillance-history experiments}
\label{appendix:mixture_history}

We conducted two complementary experiments to examine how the contribution of simulated training depends on the forecasting setting and the amount of available surveillance history. First, we varied the proportion of simulated examples during training while holding the remaining training settings fixed. Second, we evaluated each trained model using different amounts of historical context at the same forecast origins. These experiments evaluated both how overall performance depends on the quantity of synthetic training as well as how that varies based on the context length.

\subsection*{Constant-mixture training}

We trained eleven models with synthetic sampling probabilities ranging from 0 to 1 in increments of 0.1. For a model with synthetic fraction $p$, each training task was independently assigned to the simulated data pool with probability $p$ and to the observational data pool with probability $1-p$: $p \in \{0,\,0.1,\,0.2,\,\ldots,\,0.9,\,1.0\}$. The selected trajectory and forecasting task were then sampled using the procedure described in the model-training section. Thus, the mixture fraction specifies the probability of selecting a data source for each task, rather than an exact fraction of observations within every batch.

The endpoints correspond to exclusively observational and exclusively simulated task sampling during infectious-disease training. All models retained the same pretrained initialization. Architecture, optimization settings, training duration, data pools, and all other training settings were held fixed; only the synthetic sampling probability changed. This design tests how allocating a fixed training budget between simulated and observational examples affects subsequent forecast performance.

Unlike the curriculum used for the released Mantis 2 model, these sampling probabilities remained constant throughout training. The constant-mixture experiments therefore characterize the effect of training-data composition without simultaneously changing the order in which the two sources are emphasized.

\subsection*{Evaluation across datasets}

All eleven models were evaluated on all evaluation datasets using the forecast origins, inputs, and scoring procedures described in the model-evaluation section. For these experiments, we summarized WIS over the complete sixteen-week forecast horizon. Using a single horizon range simplifies comparisons across training mixtures and context lengths.

The dataset-by-mixture heatmap (Figure \ref{fig:synthetic_blends}) displays the twelve Project Tycho datasets. This restriction is for visualization only; the broader analysis includes all evaluation datasets. Displaying the Project Tycho subset keeps the heatmap readable while comparing historical U.S. disease records drawn from the same data source and broadly similar surveillance settings.

To show the relative effect of training mixture within each dataset, heatmap scores were normalized by the lowest WIS attained across the eleven mixtures. For dataset $d$ and synthetic sampling probability $p$, the displayed value is
\begin{equation*}
R_{d,p}
=
\frac{
\mathrm{WIS}_{d,p}
}{
\min_{p' \in \{0,\,0.1,\,\ldots,\,1.0\}}
\mathrm{WIS}_{d,p'}
}.
\end{equation*}
A value of 1.00 therefore denotes the best-performing mixture for that dataset, and larger values indicate worse performance relative to that minimum. This normalization makes differences among mixtures visible without allowing differences in the absolute WIS scale across datasets to mess up the heatmap.

\subsection*{Controlled variation in surveillance history}

To test whether the value of simulated training depends on the amount of historical data for a particular dataset, we evaluated every mixture model with context lengths of 8, 16, 26, 52, and 104 weeks. Eight weeks represents the minimum context considered, approximately two months. Sixteen weeks provides an intermediate context of approximately four months, while 26, 52, and 104 weeks correspond to approximately half a year, one year, and two years.

At each forecast origin, historical inputs were truncated to retain only the most recent $L$ weekly observations, where $L$ is the selected context length. The forecast origin and the subsequent sixteen-week target window were held fixed across context lengths and training mixtures. Models were evaluated with their existing weights, without retraining for a particular context length.

Holding forecast origins fixed ensures that comparisons concern the same future observations and outbreak phases. Consequently, differences between context-length conditions reflect the effect of changing the historical information available to the model, rather than evaluating short histories early in an outbreak and long histories at a different stage.

\subsection*{Improvement relative to observational training}

For each context length, mixture-model performance was compared with the exclusively observational training condition evaluated on the same forecast instances. Percentage improvement was defined as
\begin{equation*}
I_{p,L} = 100 \times \frac{\mathrm{WIS}_{0,L} - \mathrm{WIS}_{p,L} }{\mathrm{WIS}_{0,L}},
\end{equation*}
where $\mathrm{WIS}_{0,L}$ is the score of the observational-only model and $\mathrm{WIS}_{p,L}$ is the score of the model trained with synthetic sampling probability $p$, both using context length $L$. Positive values indicate that including simulated training examples improves performance; negative values indicate worse performance than observational-only training.

Comparing these improvements across context lengths tests whether additional surveillance history changes the relative benefit of simulated training. In particular, the comparison between eight and 104 weeks contrasts forecasts informed by only a short recent record with forecasts that can draw on approximately two years of local disease activity.

\section{Additional approaches considered}
\label{appendix:additional_approaches}

We investigated several additional approaches that could potentially improve forecast performance: epimodulation \cite{epimodulation}, ensembling across subsets of input variables, and joint training of multiple models using negative correlation learning \cite{ncl}. These experiments informed the design of the released model by examining both forecast accuracy and the computational and implementation costs of additional processing. To be clear: even if we did not add these methods to the final model, we are not trying to criticize their performance overall, just that we did not find them to be the most useful in our case.

\subsection*{Epimodulation}

Epimodulation augments existing forecasts with an adjustment motivated by epidemiological dynamics, particularly susceptible depletion, to improve anticipation of epidemic peaks \cite{epimodulation}. It provides a way to incorporate mechanistic information into forecasts without replacing the underlying forecasting model. We hypothesized that this adjustment might offer limited additional benefit for Mantis 2 because mechanistic simulation already exposes the model to susceptible depletion and related outbreak dynamics during training. Nevertheless, the approach was straightforward to apply and provided a useful test of whether an explicit adjustment could complement the dynamics learned through simulation. In our experiments, applying epimodulation to Mantis 2 did not improve performance and slightly increased forecast error.

To examine whether the adjustment would be more useful without infectious-disease simulation training, we also applied it to the model trained exclusively on observational disease data. This comparison was motivated by the improvements reported for empirical forecasting methods in the original epimodulation paper \cite{epimodulation}. Adjustment parameters were calibrated using only previously observed outcomes. Base and adjusted forecasts were evaluated on matched forecast instances, with identical nonnegativity and quantile-sorting procedures applied to both.

Epimodulation did not improve aggregate performance for the observational-only model (Table \ref{tab:epimodulation}). Pooled WIS increased by 1.06\%, 0.96\%, and 0.66\% over weeks 1--4, 1--8, and 1--16, respectively (lower is better). At the sixteen-week horizon range, WIS improved on four of the thirty datasets, was unchanged on one, and worsened on the remaining twenty-five. The median dataset-level change was a 2.81\% increase in WIS.

Effects nevertheless varied substantially across datasets. For example, epimodulation reduced sixteen-week WIS by 63.78\% for influenza in Austria, while increasing it by 22.41\% for dengue in Singapore (for whatever reason, the observation-only model performed very poorly on this Austrian flu data compared to all other models, including the final model, which has a WIS improvement of over 70\% relative to observation-only). These results indicate that the adjustment could be helpful in particular settings, but did not provide a consistent improvement across our evaluation.

The absence of an aggregate benefit for the observational-only model also means that redundancy with simulation training cannot, by itself, explain these findings. Differences in the underlying forecast model, training procedure, forecasting targets, horizons, and calibration procedure may affect the usefulness of the adjustment (the most probably explanation being that our models were trained on vastly more infectious disease data than the models in their paper). We therefore interpret these experiments as evidence against including our tested epimodulation procedure in the default Mantis 2 pipeline, rather than as evidence against epimodulation generally.

\begin{table}[htbp]
\centering
\caption{
Epimodulation applied to the observational-only model. WIS was calculated in $\log(1+y)$ space on matched forecast instances and pooled using the number of scored target values as weights. Positive percentage improvement favors epimodulation; negative values indicate increased WIS.
}
\label{tab:epimodulation}
\begin{tabular}{lrrr}
\toprule
Horizon range & Base WIS & Adjusted WIS
& Improvement (\%) \\
\midrule
Weeks 1--4  & 0.213999 & 0.216261 & -1.06 \\
Weeks 1--8  & 0.295455 & 0.298283 & -0.96 \\
Weeks 1--16 & 0.392063 & 0.394660 & -0.66 \\
\bottomrule
\end{tabular}
\end{table}

\subsection*{Ensembling across input-variable subsets}

We investigated whether a single trained Mantis 2 model could produce more accurate forecasts by combining predictions generated from different subsets of the available input variables. This experiment was motivated by our prior work emphasizing that infectious disease ensembles benefit from complementary predictive information, while correlated errors can limit the gains from combining forecasts \cite{ensemble_independence}. Varying the input subsets provides a way to change the information on which each forecast is conditioned while retaining the same trained model. We hypothesized that forecasts based on different combinations of surveillance signals might make sufficiently different errors to improve accuracy when combined.

For each eligible multivariate forecasting task, we generated the default forecast using all available variables and additional forecasts using subsets of those variables. We evaluated two ensemble variants: one using all nonempty subsets, including univariate inputs, and one restricted to subsets containing at least two variables (since univariate forecasts are often less accurate \cite{mantis}).

For each target, corresponding forecast quantiles were averaged with equal weight across the included subsets containing that target. Averaging was performed in $\log(1+y)$ space. All forecasts used the same trained weights, and no additional training was required. Comparisons with the default forecast used matched origins and target observations within the eligible multivariate tasks.

The improvements were small (Table \ref{tab:subset_ensembles}). Ensembling all nonempty subsets reduced pooled WIS by 0.18\% over weeks 1--16, but slightly increased WIS over weeks 1--4 and 1--8. Excluding univariate subsets produced improvements across all three horizon ranges, reducing pooled WIS by 0.15\%, 0.14\%, and 0.27\%, respectively.

Although this approach required only one set of model weights, it increased the number of forecast configurations evaluated at inference. With six available variables, there are 63 nonempty subsets, or 57 subsets when univariate inputs are excluded. These configurations can be batched, but still require substantially more inference computation than a single forecast using all variables. Given the modest aggregate gains, we retained the full-input forecast as the default to support rapid and straightforward deployment.

\begin{table}[htbp]
\centering
\caption{
Relative improvement in pooled WIS from ensembling predictions across variable subsets, compared with a single forecast using all available variables. Results are restricted to eligible multivariate tasks and use matched forecast instances. Positive values indicate lower WIS than the default forecast.
}
\label{tab:subset_ensembles}
\begin{tabular}{lrrr}
\toprule
Ensemble & Weeks 1--4 & Weeks 1--8 & Weeks 1--16 \\
\midrule
All nonempty subsets
& -0.03\% & -0.07\% & 0.18\% \\
Subsets with at least two variables
& 0.15\% & 0.14\% & 0.27\% \\
\bottomrule
\end{tabular}
\end{table}

\subsection*{Negative correlation learning}

We also investigated ensembles of multiple Mantis models trained with objectives that encouraged complementary prediction errors, following the motivation of negative correlation learning \cite{ncl}. Averaging models with similar errors provides limited benefit; encouraging less correlated or negatively correlated errors during training may allow individual models to compensate for one another when their predictions are combined.

These experiments produced forecast improvements of several percent, but required training and deploying multiple models. Compared with a single model, this increased training cost, model storage requirements, and inference computation, as well as the complexity of maintaining the forecasting pipeline.

We judged these gains insufficient to justify the additional requirements for the default release, whose design emphasizes accessibility and routine use on modest computing resources. This is a deployment trade-off rather than a conclusion that ensemble training is ineffective: applications with greater computational resources or a stronger preference for incremental accuracy may reasonably favor an ensemble.

\end{appendices}

\end{document}